\documentclass[letterpaper]{article} 
\usepackage{aaai24}  % DO NOT CHANGE THIS
\usepackage{times}  % DO NOT CHANGE THIS
\usepackage{helvet}  % DO NOT CHANGE THIS
\usepackage{courier}  % DO NOT CHANGE THIS
\usepackage[hyphens]{url}  % DO NOT CHANGE THIS
\usepackage{graphicx} % DO NOT CHANGE THIS
\usepackage{natbib}  % DO NOT CHANGE THIS AND DO NOT ADD ANY OPTIONS TO IT
\usepackage{caption} % DO NOT CHANGE THIS AND DO NOT ADD ANY OPTIONS TO IT
\usepackage{algorithm}
\usepackage{algorithmic}
\usepackage{times}
\usepackage{epsfig}
\usepackage{graphicx}
\usepackage{amsmath}
\usepackage{amssymb}
\usepackage{booktabs}
\usepackage{bbold}
\usepackage{xcolor}
\usepackage{caption}
\usepackage{multirow}
\usepackage{soul}
\usepackage{comment}
\usepackage{xcolor}
\usepackage{enumitem}
\usepackage[title]{appendix}

\usepackage{newfloat}
\usepackage{listings}
\DeclareCaptionStyle{ruled}{labelfont=normalfont,labelsep=colon,strut=off} % DO NOT CHANGE THIS
\floatstyle{ruled}
\newfloat{listing}{tb}{lst}{}
\floatname{listing}{Listing}
\title{An Image is Worth Multiple Words: Multi-attribute Inversion\\for Constrained Text-to-Image Synthesis}
\author{Aishwarya Agarwal, Srikrishna Karanam, Tripti Shukla, and Balaji Vasan Srinivasan\\
Adobe Research, Bengaluru India \\
{\tt \scalebox{.7}{\{aishagar,skaranam,trshukla,balsrini\}@adobe.com}}
}

\usepackage{bibentry}
\begin{document}

\twocolumn[{
\renewcommand\twocolumn[1][]{#1}%
\maketitle
\begin{center}
 \centering
 \captionsetup{type=figure}
 \includegraphics[width=0.9\textwidth]{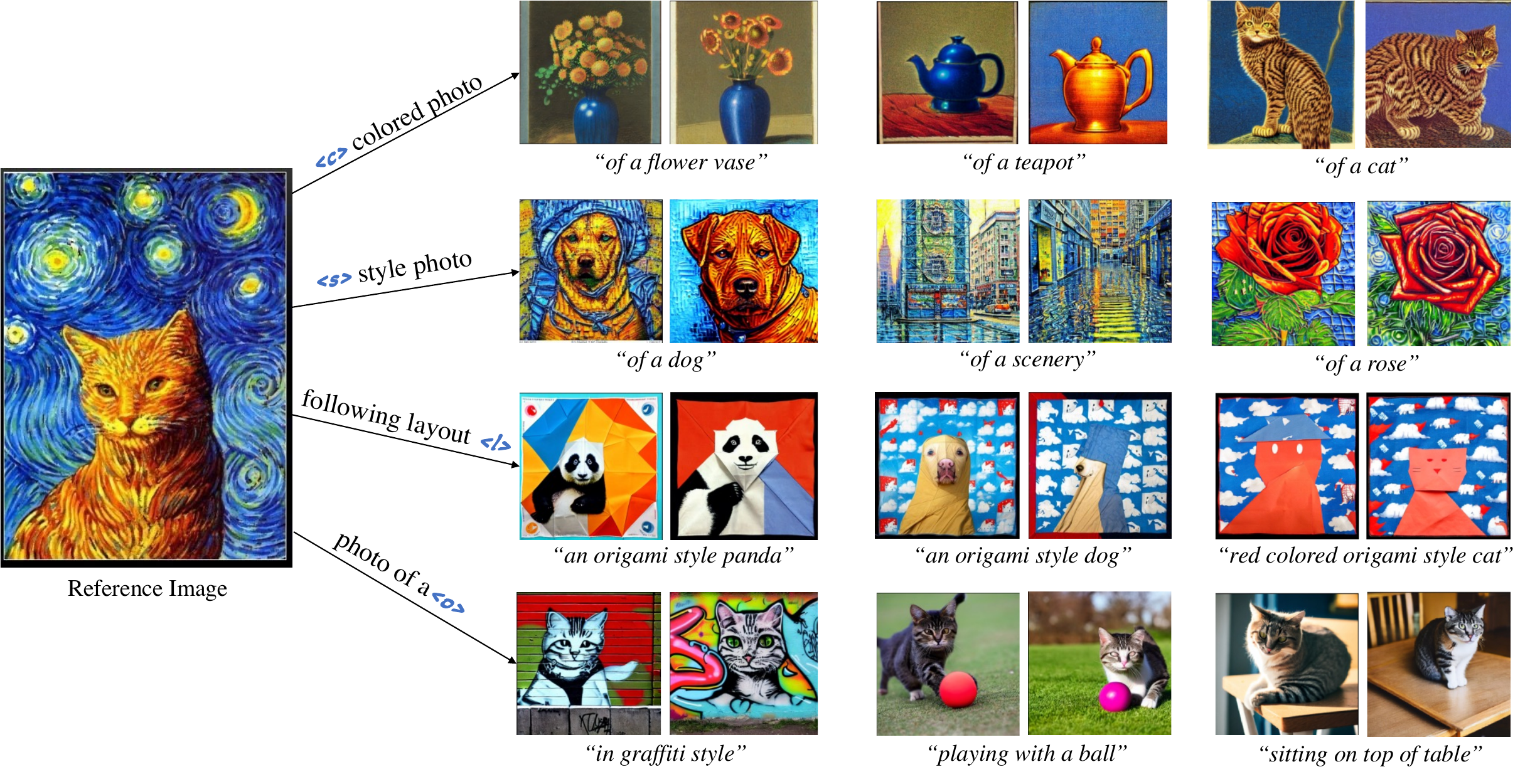}
 %\vspace{-20pt}
 \caption{Given a single reference image, we propose a new algorithm MATTE to extract four tokens, one each for color, style, object, and layout properties of the image. They can then be used for attribute-guided image synthesis as shown in columns 2-4.}
 \label{fig:teaser_qual}
\end{center}
}]
% Give a single reference image and a text prompt, our proposed method can accurately generate images constrained on multiple attributes namely color, style, object and layout of the reference image.
\begin{abstract}

We consider the problem of constraining diffusion model outputs with a user-supplied reference image. Our key objective is to extract multiple attributes (e.g., color, object, layout, style) from this single reference image, and then generate new samples and novel compositions with them. One line of existing work proposes to invert these reference images into a single textual conditioning vector, enabling generation of new samples with this learned token. These methods, however, do not learn multiple tokens that are necessarily required to condition model outputs on the multiple attributes noted above. Another line of techniques expand the inversion space to learn multiple embeddings but they do this only along the layer dimension (e.g., one per layer of the DDPM model) or the timestep dimension (one for a set of timesteps in the denoising process), leading to suboptimal attribute disentanglement.

To address the aforementioned gaps, the first contribution of this paper is an extensive analysis to determine which attributes are captured in which dimension of the denoising process. As noted above, we consider both the time-step dimension (in reverse denoising) as well as the DDPM model layer dimension. We observe that often a subset of these attributes are captured in the same set of model layers and/or across same denoising timesteps. For instance, color and style are captured across same U-Net layers, whereas layout and color are captured across same timestep stages. Consequently, an inversion process that is designed only for the time-step dimension or the layer dimension is insufficient to disentangle all attributes. This leads to our second contribution where we design a new multi-attribute inversion algorithm, \textbf{MATTE}, with associated disentanglement-enhancing regularization losses, that operates across both dimensions and explicitly leads to four disentangled tokens (color, style, layout, and object). We conduct extensive evaluations to demonstrate the effectiveness of the proposed approach.

\end{abstract}

\section{Introduction}
\label{sec:intro}

% add citations before @srikrishna gets to it

%By tapping into the capabilities of pre-trained diffusion models crafted for the purpose of generating images from text, we gain a highly adaptable technique for guiding image generation through free-form text inputs. Nowadays, diffusion models have also been adapted to allow users generate personalized images based on their visual preferences, for instance, image generation guided by visual cues from a reference image. In this paper, we consider the problem of guiding image generation by multiple attributes namely color, layout, style, and object semantics from a reference image. Our key objective is to extract attributes from this single reference image and then generate new samples as well as novel compositions with these learned attributes. One line of existing work \cite{gal2022image,ruiz2023dreambooth,kumari2023multi} inverts the reference image into a single textual conditioning vector, allowing users to generate new samples with this newly learned token. But since they do not learn multiple tokens, they are only able to condition the image generation process on the reference image as a whole and not separately on the different attributes of reference image mentioned above. 

We consider the text-to-image class of generative diffusion models \cite{ho2020denoising, rombach2022high, sohl2015deep, song2020denoising} and explore the problem of conditioning them based on a user-supplied reference image. In particular, we seek to extract multiple attributes (\texttt{color}, \texttt{object}, \texttt{style}, and \texttt{layout}) from the reference to synthesize images with any combination of these learned attributes. While there has been much work in personalizing text-to-image diffusion models \cite{gal2022image, ruiz2023dreambooth, kumari2023multi, voynov2023p+, zhang2023prospect}, they lack explicit control on which attributes from the reference are to be reflected in the model outputs. For instance, one line of recent work \cite{gal2022image,ruiz2023dreambooth,kumari2023multi} inverted the reference image into a single learned token and used it to synthesize new samples. Since this process learned only a single conditioning vector, it was not able to disentangle the inherently multi-attribute information from the reference image (e.g., \texttt{color}, \texttt{object}, \texttt{layout}, and \texttt{style} all constitute complementary and separate pieces of information). Consequently, these techniques only generate more samples ``like" the reference but fail if we seek more control (e.g., in synthesizing samples that follow the \texttt{color}, or \texttt{style}, or \texttt{layout} and \texttt{object} of the reference, etc.).

%Recently there have been another line of works \cite{voynov2023p+, zhang2023prospect} that expand the inversion space to learn multiple textual conditioning vectors in an attempt to guide the image generation process using multiple attributes of the reference image. One such work P+ \cite{voynov2023p+} derives these tokens by inverting the reference image along the cross-attention layers of U-Net whereas another work ProSpect \cite{zhang2023prospect} inverts the reference image along the denoising timesteps dimension. P+ and Prospect attempted to analyse attribute distribution of diffusion models across layer and timestep dimension respectively but their analysis is not exhaustive for the reasons mentioned below.

There have also been attempts to expand the inversion space by learning multiple token embeddings. For instance, P+ \cite{voynov2023p+} learned these tokens by inverting the reference image along all the 16 cross-attention layers of the DDPM model \cite{ronneberger2015u}, whereas ProSpect \cite{zhang2023prospect} inverted the reference image along the denoising timestep dimension by dividing the steps into 10 stages. As we discuss next, even these strategies are insufficient to disentangle all attributes.

\begin{figure}[h]
    \centering
    \includegraphics[width = 0.8\linewidth]{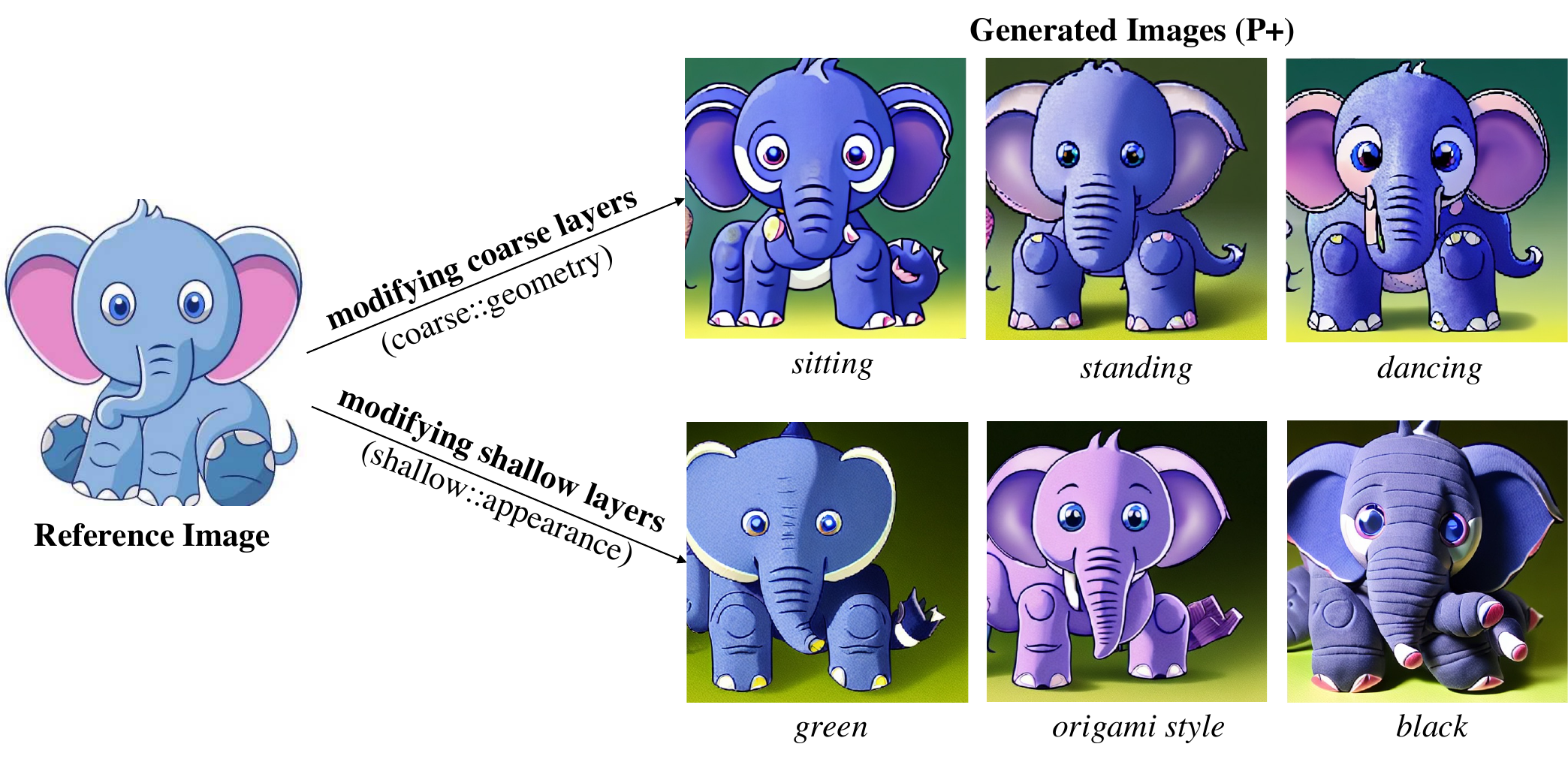}
    % \vspace{-18pt}
    \caption{Attribute entanglement in P+ \cite{voynov2023p+}.}
    % \vspace{-14pt}
    \label{fig:Pplus_limitn}
\end{figure}

First, the key findings from P+ \cite{voynov2023p+} were that semantic information is captured in coarse layers of the DDPM model (U-Net \cite{ronneberger2015u}) whereas appearance information (e.g., color) is captured in the shallow layers. However, as we will show in our work, \texttt{layout} and \texttt{object} semantics are captured in the same set of coarse layers of the DDPM model whereas \texttt{color} and \texttt{style} share the same set of shallow layers. This suggests that the attempted disentanglement in P+ \cite{voynov2023p+} by learning multiple tokens across the DDPM layer dimension is insufficient because such an inversion space will not disentangle \texttt{color} from \texttt{style} and \texttt{layout} from \texttt{object}. To understand this clearly, consider the example shown in Figure~\ref{fig:Pplus_limitn}. Here, we invert the elephant reference image using the P+ \cite{voynov2023p+} technique and synthesize new images by modifying either of the coarse or shallow layers. The synthesized images in the first row show that they have \textit{both} \texttt{object} and \texttt{layout} semantics from the reference image and cannot be conditioned on either of them individually since they are captured in the same set of DDPM model layers. Similar observations hold for the \texttt{color} and \texttt{style} pair as well since they are captured in the same set of layers (see second row in Figure~\ref{fig:Pplus_limitn}).

%On the other hand, Prospect analyses the attribute distribution across the denoising timestep dimension and concludes that attributes like color and layout are captured in the initial denoising timesteps, object semantics towards the middle and fine-grained details towards the end. As we will show in our analysis, stylistic features also happen to be captured in the initial denoising timesteps along with color and layout, thereby making it difficult to disentangle the three attributes layout, color and style across the timestep dimension. See the example in Figure~\ref{fig:prospect_limitation}, where we use the reference image of a \textit{pinkish blue cartoon style elephant} as a layout reference for text-to-image generation. As we can see, along with layout, the newly generated images happen to follow the same color and style as well from the reference image.

\begin{figure}[h]
    \centering
    \includegraphics[width = 0.8\linewidth]{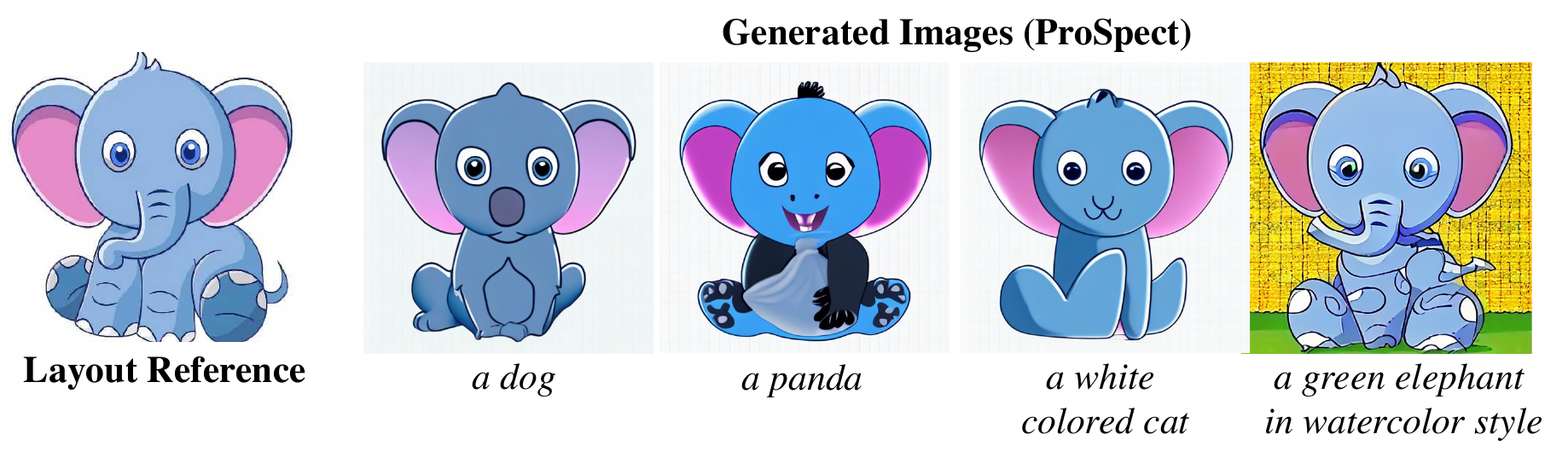}
    %\vspace{-18pt}
    \caption{Attribute entanglement in ProSpect.}
    %\vspace{-14pt}
    \label{fig:prospect_limitation}
\end{figure}

Next, Prospect \cite{zhang2023prospect} proposed to divide the denoising process into several stages (each comprising a few timesteps) and concluded that attributes like \texttt{color} and \texttt{layout} are captured in the beginning (first several timesteps), \texttt{object} semantics towards the middle, and fine-grained details towards the end of the denoising process. As we will show in our work, the \texttt{style} attribute is also captured in the first few timesteps, thereby making it challenging to disentangle \texttt{color}, \texttt{layout}, and \texttt{style} if we consider only this timestep dimension when learning these tokens. To understand this clearly, consider the example in Figure~\ref{fig:prospect_limitation}, where we use the reference image (first column) of a \textit{pinkish blue cartoon style elephant} for text-to-image generation. In the synthesized images (columns 2-5), one can note they follow not only the \texttt{layout} but also the \texttt{color} and \texttt{style} information from the reference image, suggesting no disentanglement among these attributes has happened after learning tokens with Prospect \cite{zhang2023prospect}. 

%We believe that an inversion space focused along just the layer dimension or the denoising timesteps dimension is insufficient for constraining text-to-image generation on a reference image in a disentangled fashion across all four attributes namely layout, color, style and object semantics for the reasons explained above, and hence there is a need to consider the two dimensions in a joint fashion. In this paper, we first conduct a thorough analysis on how joint textual-conditioning across layer and timestep dimension influences the generation process of diffusion models extensively for all four attributes. 

\begin{figure}[h]
    \centering
    \includegraphics[width = \linewidth]{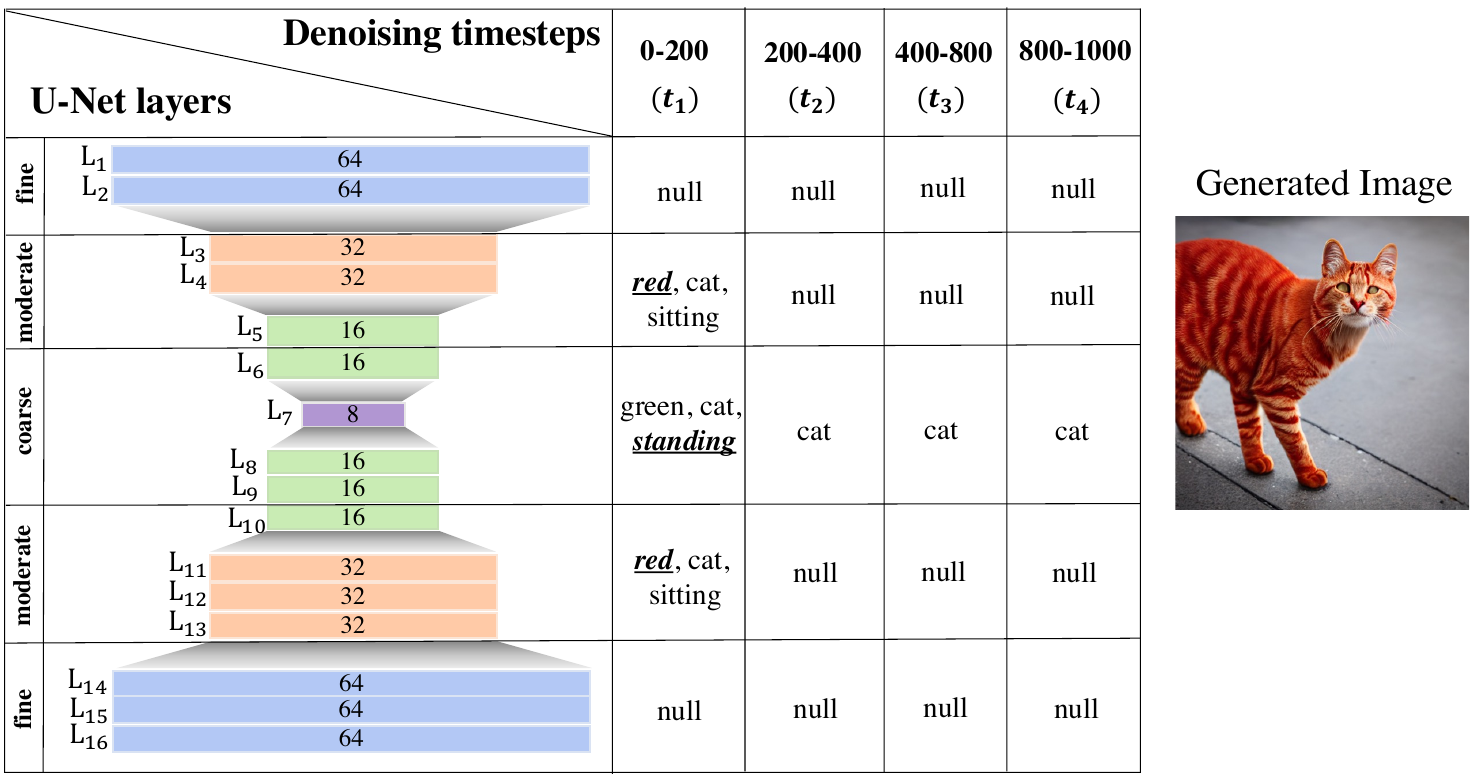}
    %\vspace{-18pt}
    \caption{Layout-Color disentanglement.}
    %\vspace{-5pt}
    \label{fig:analysis_layout_color}
\end{figure}

Based on the aforementioned observations and limitations of P+ \cite{voynov2023p+} and Prospect \cite{zhang2023prospect}, the key insight of this paper is that an inversion strategy solely focused on either the layer dimension or the denoising timestep dimension is insufficient to disentangle all our attributes of interest from a reference image and use them for controlled text-to-image synthesis. In fact, we need to consider both dimensions \textit{jointly} when designing an inversion strategy so as to learn meaningful individual attribute tokens for subsequent synthesis. To this end, the key contributions of this paper are two-fold. First, we conduct an extensive and exhaustive layer-and-timestep analysis to determine which layers and which timesteps influence what attributes during the generation process. While we discuss complete details in Section~\ref{sec:analysis}, we present some sample results here. In Figure~\ref{fig:analysis_layout_color}, we show all the 16 cross-attention layers of the DDPM U-Net model \cite{ronneberger2015u} as well as four different stages ($t_{1}-t_{4}$) of the denoising process. Here, based on the final generated image (a red standing cat), one can note that the text conditions corresponding to the color \texttt{red} were specified in the $L_3$ - $L_5$ $\&$ $L_{10}$ - $L_{13}$ layers (see first column, $t_1$ stage) and had the most impact on the final image. For instance, despite the input \texttt{green} in $L_6$ - $L_9$ layers, the final image has a red cat. Similarly, despite specifying the layout \texttt{sitting} in layers $L_3$ - $L_5$ $\&$ $L_{10}$ - $L_{13}$, the final generation only respected \texttt{standing} that was provided to $L_6$ - $L_9$. This shows that while \texttt{color} and \texttt{layout} are captured along the same timesteps (from Prospect \cite{zhang2023prospect}), they can be disentangled along the layer dimension ($L_3$ - $L_5$ $\&$ $L_{10}$ - $L_{13}$ for color and $L_6$ - $L_9$ for layout).

Informed by results like those above, our second contribution is a new \textbf{\underline{M}}ulti-\textbf{\underline{Att}}ribute Inv\textbf{\underline{e}}rsion algorithm called \textbf{MATTE} that is explicitly designed to consider both the DDPM model layer and the denoising timestep dimension as part of the learning process. Specifically, we propose four new learnable tokens, one for each of \texttt{color}, \texttt{object}, \texttt{layout}, and \texttt{style}, and design our learning objectives so that each of the four tokens are trained to influence either separate layers in the model or separate timesteps in the denoising process. Figure~\ref{fig:teaser_qual} shows our results with the individual $<c>$, $<o>$, $<s>$, and $<l>$ tokens. Using them as part of new prompts leads to meaningful results (e.g., flower vase/teapot/cat in first row follow the color $<c>$ of the reference, the dog/scenery/rose in second row follow the style $<s>$ of the reference, we get cat images with object $<o>$ as desired in the last row and so on).

To summarize, the key contributions of this paper are:

\begin{itemize}
    %\item We present an extensive analysis to determine which attributes are captured in which dimension of the denoising process where we consider both the time-step dimension (in reverse denoising) as well as the DDPM model layer dimension. This helps us determine key reasons behind imperfect disentangelemt of existing works as shown in Figure~\ref{,} [refer P+ and prospect limitation figs]
    \item We present an extensive multi-attribute disentanglement analysis jointly across both the DDPM model layer dimension and the reverse denoising timestep dimension to understand which attributes are captured at which stage and along which dimension during the generation process, thereby discovering reasons why existing inversion algorithms fail as shown in Figures~\ref{fig:Pplus_limitn} and~\ref{fig:prospect_limitation}.
    %\item We propose a novel inversion technique, X that inverts reference images into a disentangled set of tokens, allowing seamless attribute-guided personalized content generation
    \item We present a novel multi-attribute inversion algorithm with four learnable tokens (for color, object, layout, and style attributes) and a principled approach to disentangling them, allowing for attribute-guided synthesis based on reference images.
    %\item The two attributes color and style are both captured during initial denoising timesteps across the moderate U-Net layers. Hence, we propose a  novel disentanglement-enhancing loss during the inversion process to ensure disentanglement of these attributes
    %\item something on analysing clip-space proximity for the learned tokens
\end{itemize}

% we'll basically show that disentangling fully across one dimension is not possible due to so and so reasons, but if you invert across both jointly, it is possible to disentangle.

% generative models also allow for personalization
% brief about some works that allow for personalization
% talk about P+ and Prospect and point to the images that highlight their limitation
% Talk about the brief observations from the analysis we'll present and connect that to the limitations highlighted for P+ and Prospect
% Then comes our proposal from technique perspective: we propose a novel textual inversion method called Prompt Spectrum Inversion (ProSpect), which learns four token embeddings in the disentangled attribute space (get name for your new space) from a single image
% then talk about the applicability and effectiveness of Prospect while pointing to some results you'll have say in the teaser
% summarize the contributions

\begin{figure*}[!ht]
    \centering
    \includegraphics[width = 0.9\linewidth]{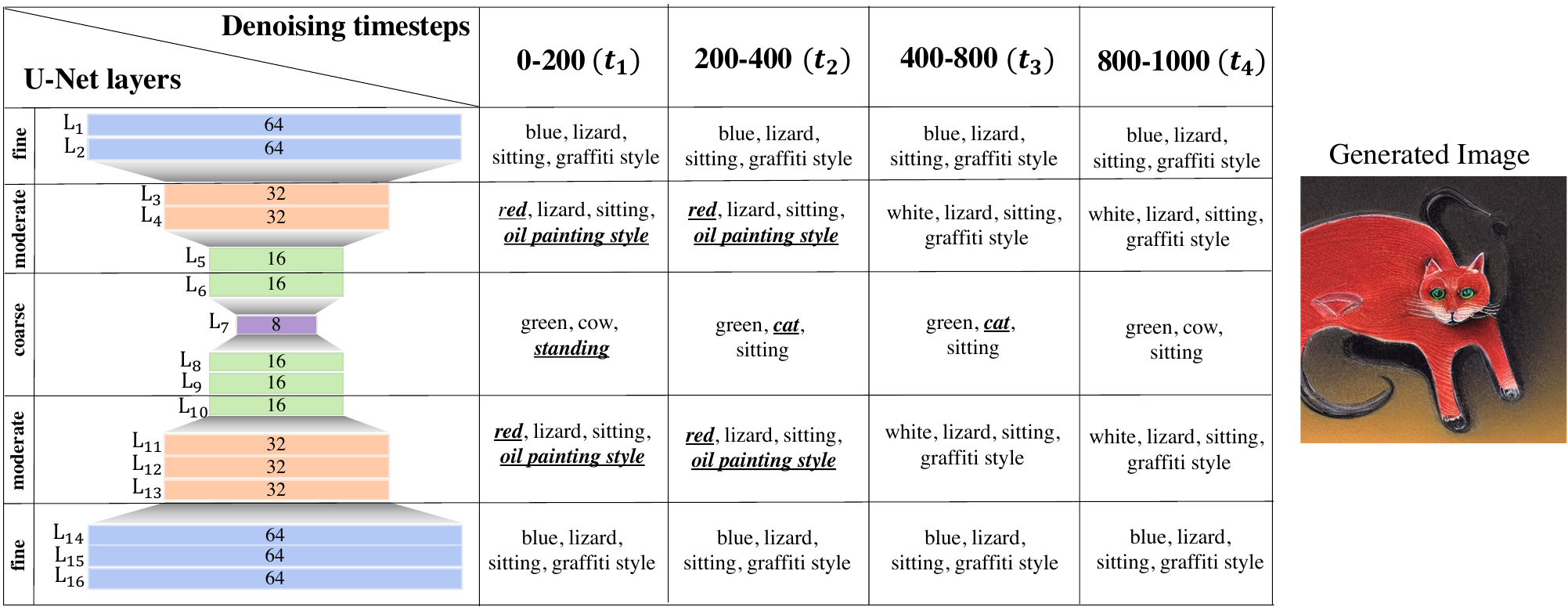}
    %\vspace{-18pt}
    \caption{Multi-prompt conditioning across U-Net layers and denoising timesteps jointly.}
    %\vspace{-14pt}
    \label{fig:analysis_overall_joint}
\end{figure*}

\section{Related Work}

Since the emergence of large-scale diffusion models for conditional image synthesis \cite{nichol2021glide, rombach2022high, ramesh2022hierarchical, saharia2022photorealistic}, there has been much recent work in adapting these models with a variety of conditioning types  \cite{zhang2023adding,mou2023t2i}. One particular line of work has been personalized inversion where a reference image is used to learn conditioning vectors to be used with novel prompts. Gal et al. \cite{gal2022image} proposed a first baseline version that learned a single vector for a new token (while keeping diffusion model weights fixed), which could then used to synthesize novel personalized variations of the reference image. On the other hand, methods like Dreambooth \cite{ruiz2023dreambooth}, CustomDiffusion \cite{kumari2023multi}, and Perfusion \cite{tewel2023key} involved finetuning the weights of the diffusion model. Inversion has been explored in GANs too \cite{bermano2022state, creswell2018inverting, lipton2017precise, xia2022gan} with both latent optimization \cite{abdal2019image2stylegan, abdal2020image2stylegan++} and model finetuning methods \cite{alaluf2022hyperstyle, roich2022pivotal, nitzan2022mystyle}. 

Our closest baselines include P+ \cite{voynov2023p+} and Prospect \cite{zhang2023prospect}, both of with enhanced the inversion space by learning more than one conditioning vector. However, they are unable to disentangle all our attributes of interest. In P+ \cite{voynov2023p+}, while tokens are learned by inverting the reference image along all the 16 cross-attention layers of the DDPM model, disentanglement of color from style and layout from object is impossible because the layout/object pair and the color/style pair are captured in the same set of layers of the DDPM model. On the other hand, in Prospect \cite{zhang2023prospect}, while tokens are learned by dividing the denoising process into multiple timestep stages, it is not able to disentangle color, layout, and object since they are all captured in the same timestep stages. To address these issues, our proposed inversion algorithm introduces new per-attribute tokens and optimizes them jointly across both layer and timestep dimension, enabling multi-attribute extraction from a reference image and synthesizing novel compositions using them.

\section{Approach}

\subsection{Layer-timestep Attribute Disentanglement}
\label{sec:analysis}
% some word that includes both dimension and attribute
% start with some lines of P+ and Prospect and summarise their main conclusions; why is it insufficient for us?
% and then comes the main figure - that covers both dimension analysis and layer analysis; and summarise the key findings
% and then we go into individual attributes
% last para: summarise and lead to inversion algorithm - has to respect the fact that there is timestep and layer conditioning jointly

%To illustrate our motivation, we start by analyzing the attribute distribution of diffusion models using text-guided image generation results. We consider both the timestep dimension (in the backward denoising) as well as the layer dimension (in the DDPM model) jointly for this analysis. 

Our first contribution is an extensive analysis of text-to-image diffusion models to understand which layers (in the DDPM model) and timesteps (in backward process) jointly are responsible for capturing attributes (we consider \texttt{color}, \texttt{style}, \texttt{layout}, \texttt{object}) during generation. As discussed in Section~\ref{sec:intro} previously, either P+ \cite{voynov2023p+} or Prospect \cite{zhang2023prospect} are unable to disentangle all attributes. To understand why and how that leads to our method in Section~\ref{sec:inversion_method}, we analyze the attribute distribution during the generation process jointly across both the layer and timestep dimensions. To this end, let us consider the example shown in Figure~\ref{fig:analysis_overall_joint}. The U-Net \cite{ronneberger2015u} that is used in the DDPM model of Stable Diffusion \cite{rombach2022high} comprises 16 cross-attention layers of resolutions 8, 16, 32, and 64 (see the figure for per-layer resolution). We partition them into three sets: coarse ($L_6$ - $L_9$), moderate ($L_3$ - $L_5$ $\&$ $L_{10}$ - $L_{13}$), and fine ($L_1$ - $L_2$ $\&$ $L_{14}$ - $L_{16}$). Similarly, we partition the denoising timesteps into four stages: $t_1$, $t_2$, $t_3$ and $t_4$. We seek to understand the timesteps and layers where the four attributes are captured during the generation process. To do this, we propose to add/remove conditioning from both timesteps and layers and analyze the output.

%For instance, see Figure~\ref{fig:analysis_overall_joint} where we show results for a case of joint prompting across U-Net layers and denoising stages. Here, for the final generated image (\textit{a red standing cat in oil painting style}), one can note that the textual conditionings corresponding to each of these attributes namely \texttt{red}, \texttt{standing}, \texttt{cat}, and \texttt{oil painting} were specified only across a subset of layers, and only along specific timesteps. In fact, despite specifying attributes like \texttt{blue} color in $L_1$ - $L_2$ $\&$ $L_{14}$ - $L_{16}$, we still see a \texttt{red} colored cat,  which clearly indicates the existence of some patterns in how these attributes are distributed across different U-Net layers and denoising stages. We will next analyse and understand the distribution of each of these attributes in depth.

\begin{figure}
    \centering
    \includegraphics[width = 0.7\linewidth]{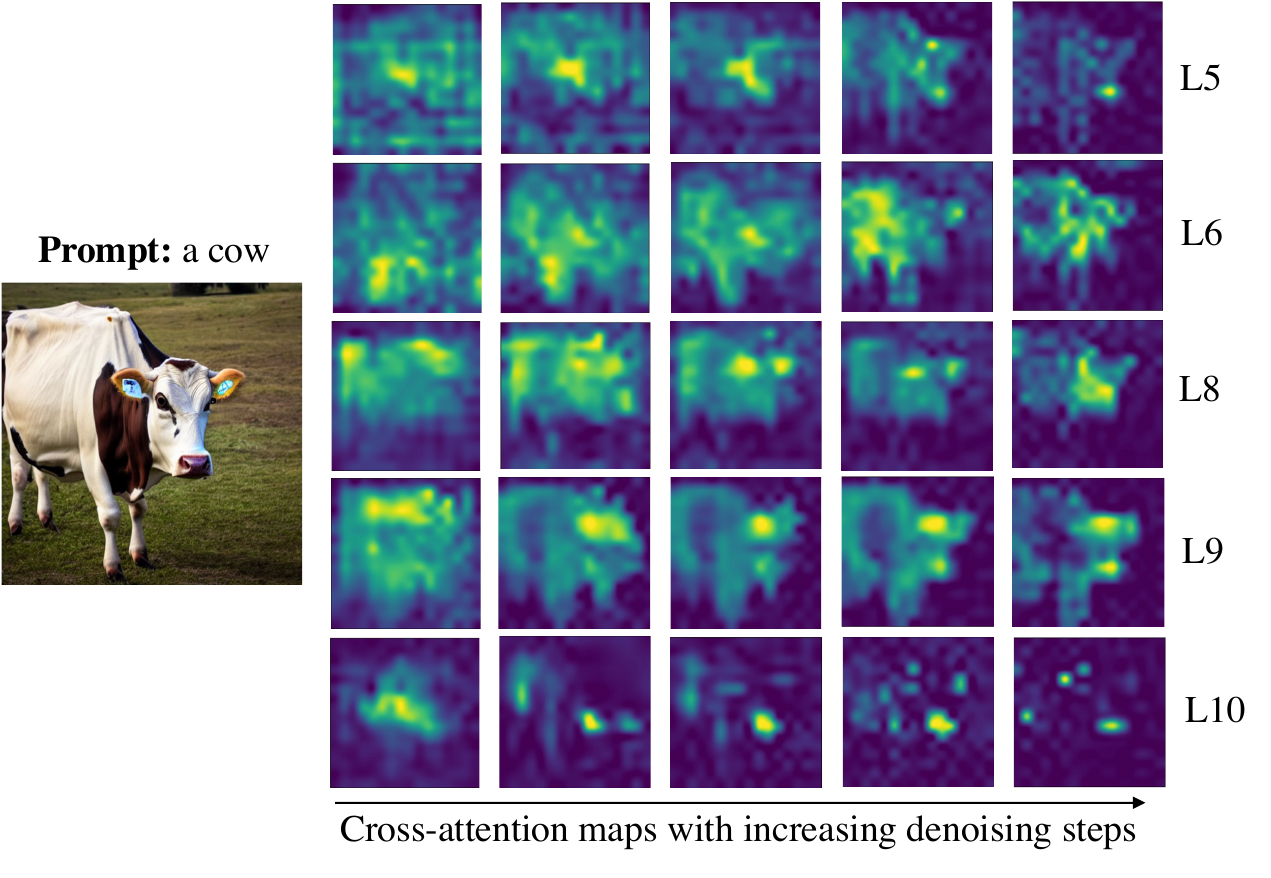}
    %\vspace{-10pt}
    \caption{Cross-attention maps for analyzing layout.}
    %\vspace{-14pt}
    \label{fig:layout_analysis}
\end{figure}

In Figure~\ref{fig:analysis_overall_joint}, we show results for one case of joint prompting across both layers and the denoising stages. For the final generated image (\textit{a red standing cat in oil painting style}), one can note that the textual conditionings corresponding to each of key attributes in the prompt (\texttt{red}, \texttt{standing}, \texttt{cat}, and \texttt{oil painting}) were specified only across a subset of layers and only along specific timesteps. In fact, despite specifying \texttt{blue} in $L_1$ - $L_2$ $\&$ $L_{14}$ - $L_{16}$ layers, we still see a \texttt{red} colored cat, suggesting the existence of some patterns in how these attributes are distributed across layers and time stages. We next discuss them:

\begin{itemize}
    \item \textbf{\textit{Color:}} Specifying colors like \texttt{green} and \texttt{blue} in conditioning for fine ($L_1$-$L_2$ $\&$ $L_{14}$-$L_{16}$) and coarse layers ($L_6$-$L_9$) respectively has no impact on the generated image (which is \texttt{red}). Similarly, colors like \texttt{white} in the later denoising stages ($t_3$, $t_4$) of moderate layers ($L_3$ - $L_5$ $\&$ $L_{10}$ - $L_{13}$) has no impact on the final generation and we indeed get a \texttt{red} cat. This indicates that \texttt{color} is captured in the initial denoising stages ($t_1$, $t_2$) across the moderate layers ($L_3$ - $L_5$ $\&$ $L_{10}$ - $L_{13}$). 

    \item \textbf{\textit{Style:}} This is similar to color. Specifiying \texttt{graffiti} and \texttt{watercolor} styles across coarse ($L_6$-$L_9$) and fine ($L_1$-$L_2$ $\&$ $L_{14}$-$L_{16}$) layers, and \texttt{graffiti} towards the later denoising stages ($t_3$, $t_4$) has no impact on the generated image. The image is still based on \texttt{oil painting} from ($t_1$, $t_2$) across ($L_3$ - $L_5$ $\&$ $L_{10}$ - $L_{13}$).

    \item \textbf{\textit{Object:}} A \texttt{cat} is generated despite specifying \texttt{cow} in the initial and later stages ($t_1$, $t_4$), suggesting \texttt{object} is captured in the middle stages ($t_2$, $t_3$). Coarse layers ($L_6$-$L_9$) seem to be responsible because specifying other types like \texttt{lizard} in other layers have no impact.

     \item \textbf{\textit{Layout:}} We can see that changing the \texttt{layout} aspects from \texttt{standing} to \texttt{sitting} after the first denoising stage has no impact on the posture of the cat being generated. This indicates \texttt{layout} is captured in the initial stages ($t_1$). Moreover, only the layers with resolution 16 are responsible. In particular, based on the per-layer cross-attention maps across timesteps in Figure~\ref{fig:layout_analysis}, one can note layout properties are predominantly captured across the initial few timesteps in layers $L_6$, $L_8$, and $L_9$.
\end{itemize}

%We consider example in Figure~\ref{fig:layout_analysis} to dive deeper into the attribute \texttt{layout} and its' behaviour along the dimensions of both U-Net layers as well as denoising timesteps to further strengthen the observations made above. The figure shows a generated image of \textit{a cow} and the corresponding cross-attention maps for all the 5 layers having resolution 16 across increasing denoising timesteps. One can note that layout gets captured majorly across the initial few timesteps in layers $L_6$, $L_8$, and $L_9$ and stays largely constant towards the later denoising timesteps.

To summarize, fine layers ($L_1$ - $L_2$ $\&$ $L_{14}$ - $L_{16}$) and the stage $t_4$ have no impact on any of the four attributes. \texttt{Color} and \texttt{style} are both captured in the initial denoising stages ($t_1, t_2$) and across moderate U-Net layers ($L_3$ - $L_5$ $\&$ $L_{10}$ - $L_{13}$). \texttt{Object} semantics are captured along the middle denoising stages ($t_2, t_3$) and across the coarse U-Net layers ($L_6$ - $L_9$). Finally, \texttt{layout} is captured in the very initial denoising stage ($t_1$) across coarse layers ($L_6$ - $L_9$). 

\subsection{MATTE: \underline{M}ulti-\underline{Att}ribute Inv\underline{e}rsion}
\label{sec:inversion_method}

Given a reference image and a base text-to-image diffusion model, the second contribution of this paper is MATTE, a new inversion algorithm that extracts a set of four tokens $<c>$, $<o>$, $<s>$, and $<l>$ for the \texttt{color}, \texttt{object}, \texttt{style}, and \texttt{layout} attributes respectively. Our algorithm design is motivated by the conclusions in Section~\ref{sec:analysis} and explicitly considers both the DDPM model layer and the timestep dimension jointly as part of the token learning process. This essentially means that the textual condition vectors in our algorithm vary across both U-Net layer dimension as well as the timestep dimension. Note that this is different from P+ \cite{voynov2023p+} where these vectors were different only for the U-Net layers and Prospect \cite{zhang2023prospect} where these vectors were different only for different timesteps of the reverse denoising process. As summarized in Section~\ref{sec:analysis}, our key insight is that we can disentangle all the four attributes only when we consider both layer and timestep dimension jointly as part of the learning process.

To this end, we divide the U-Net into \texttt{coarse}, \texttt{moderate} and \texttt{fine} layers and the forward diffusion process of 1000 steps into four stages $t_1'$ (800-1000), $t_2'$ (600-800), $t_3'$ (200-600), $t_4'$ (0-200). Note that $t_1', t_2', t_3', t_4'$ in forward diffusion corresponds to the $t_1, t_2, t_3, t_4$ (notation used in Section~\ref{sec:analysis}) respectively of the backward denoising process. Hence, properties of backward denoising stages translate directly to the corresponding forward diffusion stages.
\begin{figure}
    \centering
    \includegraphics[width = 0.85\linewidth]{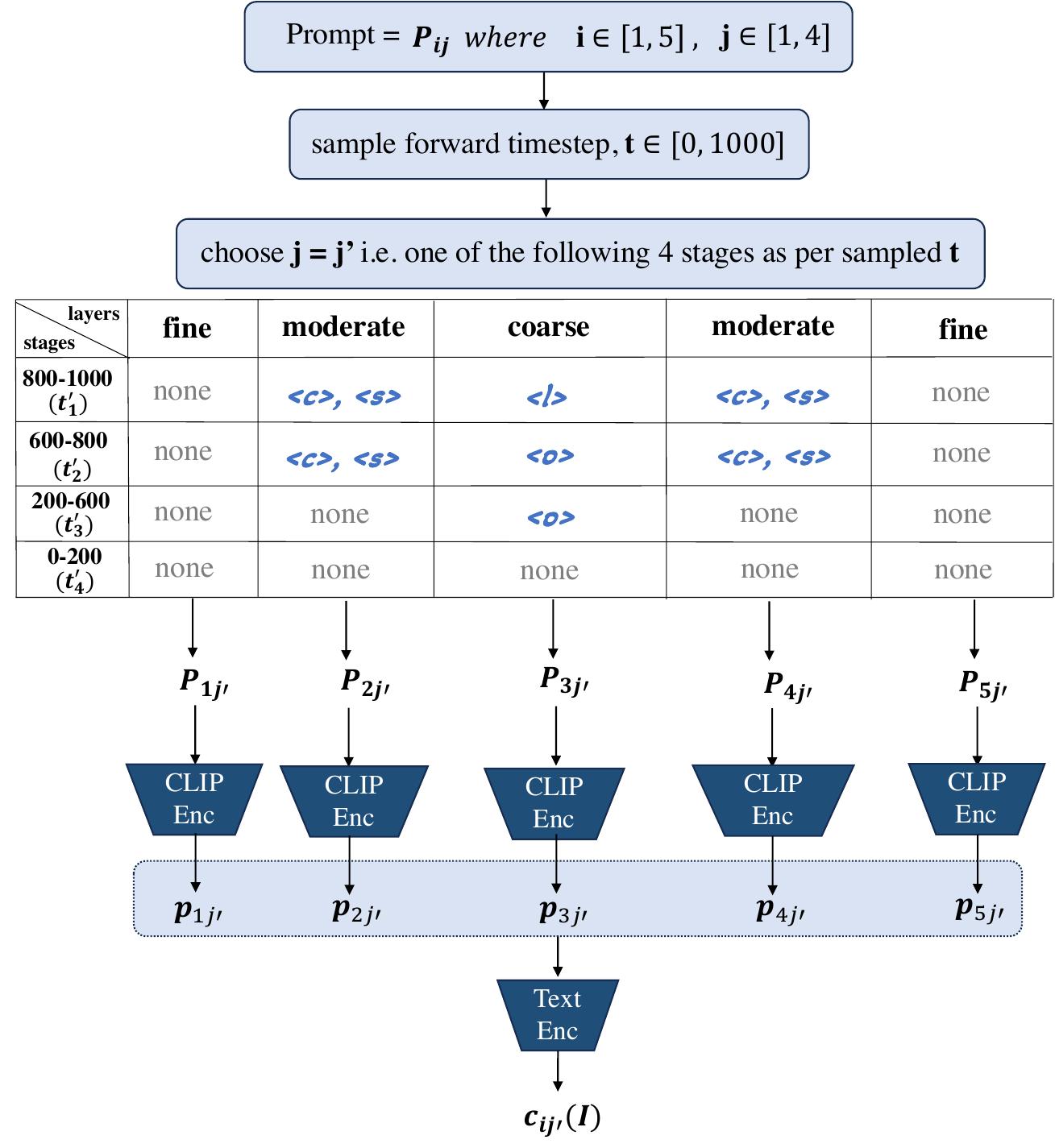}
    %\vspace{-18pt}
    \caption{Computing the input conditioning in MATTE.
    %Pipeline for computation of input textual conditioning for the diffusion model in proposed X inversion.
    }
    %\vspace{-14pt}
    \label{fig:method_tokens}
\end{figure}
We summarize the salient features of our algorithm in Figure~\ref{fig:method_tokens}. We use $P_{ij}$ to specify how the input prompt translates to the conditioning vector. The $i \in [1, 5]$ corresponds to the five layer subsets (shown in Figure~\ref{fig:method_tokens}) whereas $j \in [1,4]$ corresponds to the four timestep stages. Consequently, $P_{ij}$ comprises of a set 4 different prompts, one for each timestep stage, where each of further comprises 5 prompts for conditioning each layer subset differently. Figure ~\ref{fig:method_tokens} also shows how our learnable tokens $<c>$, $<o>$, $<s>$, and $<l>$ are part of the input prompt across the various layers and timestep stages. For instance, as noted in Section \ref{sec:analysis}, the \texttt{object} feature get captured only in the coarse layers ($L_6$ - $L_9$) and the middle $t_2, t_3$ backward denoising stages. Consequently, one can note the token $<o>$ is explicitly designed to condition only the coarse layers and the forward  $t_2', t_3'$ diffusion stages. This means if the sampled timestep $t \in t_3'$ 
% \snote{need to clean up this time notation, we've used both t2-t3 and t3 in various places to refer to a ``stage"} 
in the forward diffusion process, only then will $<o>$ end up influencing the final conditioning vector across the coarse U-Net layers. Similar observations can be derived from Figure~\ref{fig:method_tokens} for $<c>$, $<s>$, and $<l>$ tokens. Subsequently, during inversion, for a particular backward pass, depending on the sampled timestep, the embeddings corresponding to only active tokens from Figure~\ref{fig:method_tokens} end up being optimized.

%As can be noted from our analysis in Section \ref{sec:analysis}, the attributes for tokens $<c>$ and $<s>$ happen to be captured across the same U-Net layers and same timestep stages. Hence we propose an additional content-style disentanglement-enhancement loss, $\mathcal{L_{CS}}$ which is added to $\mathcal{L_D}$ described in Eq. 2 above as part of the inversion process.
%\begin{equation}
 %   \mathcal{L_{CS}} = \|\textbf{c} - \textbf{s}\|_2^2 -  \|\textbf{c}_{gt} - \textbf{s}\|_2^2, 
%\end{equation}
%where \textbf{c} refers to the encoded vector for token $<c>$, \textbf{s} refers to the encoded vector for any randomly chosen style, and $\textbf{c}_{gt}$ refers to the embedding for a concatenation of ground truth colors present in the reference image. These ground truth colors can be extracted using any off-the-shelf color palette extraction methods like Color Thief \cite{dhakar2015color}. The intuition here is to push the embedding being learned for token $<c>$ closer to the space of colors in clip space without putting hard constraints on the embedding.

MATTE's learning objective comprises three parts. The first part is the standard reconstruction loss: 
\begin{equation}
    \mathcal{L_R} = \mathbb{E}_{z \sim E(I),t,p, \epsilon \sim \mathcal{N}(0,1)}[\|\epsilon - \epsilon_{\Theta}(z_{t}, t, p_{j})\|_2^2], 
\end{equation}
where $p_{j}$  comprises learnable embeddings for a subset of the $<c>$, $<o>$, $<s>$, and $<l>$ tokens depending on the sampled timestep t $\in$ [0, 1000] in the forward diffusion process. Next, \texttt{color} and \texttt{style} attributes are captured across the same layers and the timestep stages (see also Figure~\ref{fig:method_tokens}). Consequently, we propose an additional color-style disentanglement loss to help disentangle these tokens:

\begin{equation}
    \mathcal{L_{CS}} = \|\textbf{c} - \textbf{s}\|_2^2 -  \|\textbf{c}_{gt} - \textbf{s}\|_2^2, 
\end{equation}
where \textbf{c} is the encoded vector of token $<c>$, \textbf{s} is the encoded vector for any randomly chosen style from a set of 30 styles like watercolor, graffiti, oil painting and so on (see supp. for full set of styles)
, and $\textbf{c}_{gt}$ is the CLIP \cite{radford2021learning} embedding of all the ground truth colors in the reference image (which we extract using the Color Thief \cite{dhakar2015color} library). 
% \snote{need a couple lines here on the set of styles from which we sample s}.
Our intuition is to push the learned embedding \textbf{c} for $<c>$ close to $\textbf{c}_{gt}$ by ensuring both are equally distant from $\textbf{s}$. This process naturally pushes $<c>$'s embedding to be close to the CLIP feature space of colors (and hence different from the embedding for $<s>$). %\snote{I feel like while $<c>$ being close to CLIP color space is coming out well, the disentanglement with $<s>$ is not super obvious- need to add a few more clear lines}.
% Rough ideas: We do it in a way that the distance between ground truth color vector and a particular style is retained in the learned color vector and that style too. This makes sure two things: (1) c being learned remains closer to the color space and respects the distance that color and style features have in the original clip space, (2) since the proposed loss is not a direct hard constraints like an MSE between c and c_gt, the CS loss ensures that the leanred c moves closer to the groun truth colors in the clip space, but also provides it enough freedom to also more from the reconstruction loss for more specific details.

%In order to further strengthen the disentanglement of $<o>$ and $<l>$, we add another loss $\mathcal{L_{O}} = \|\textbf{o} - \textbf{o}_{gt}\|_2^2$, where \textbf{o} refers to the encoded vector for token $<o>$ and $\textbf{o}_{gt}$ refers to the ground truth class information of the reference image. This ground truth can either be user-provided or can be obtained by nearest neighbour retrieval for clip image embedding of the reference image.  
Finally, from Section~\ref{sec:analysis} and Figure~\ref{fig:method_tokens}, \texttt{object} and \texttt{layout} inform the same set of coarse U-Net layers. To further disentangle $<o>$ and $<l>$, we propose a regularization on the learned token for $<o>$ by ensuring it respects the class of the object depicted in the reference image. We do this by computing the ground-truth class label's CLIP vector and enforcing it to be close to $<o>$'s vector:  

\begin{equation}
    \mathcal{L_{O}} = \|\textbf{o} - \textbf{o}_{gt}\|_2^2
\end{equation}
where \textbf{o} is the learned vector for token $<o>$ and $\textbf{o}_{gt}$ is the ground truth. %The ground truth can either be user-provided or can be obtained by nearest neighbour retrieval for clip image embedding of the reference image. 
MATTE's overall loss function is:
\begin{equation}
    \mathcal{L}_{inv} = \mathcal{L_R} + \lambda_{\mathcal{CS}}\mathcal{L_{CS}} + \lambda_{\mathcal{O}}\mathcal{L_O} 
    \label{eq:proposed}
\end{equation}
where $\lambda_{CS} = \lambda_{O} = 0.1$

\begin{figure*}[!ht]
    \centering
    \includegraphics[width = 0.9\linewidth]{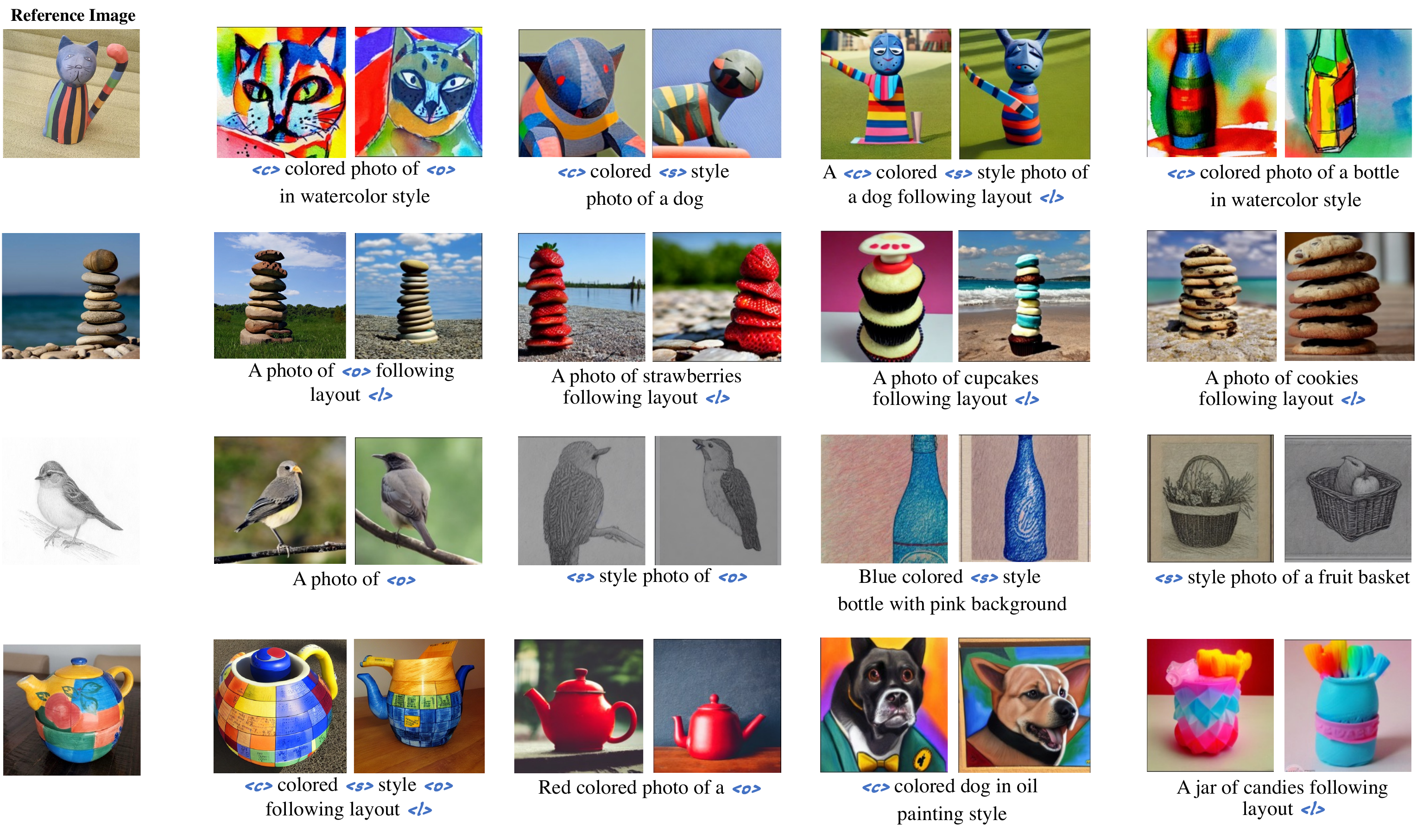}
    %\vspace{-14pt}
    \caption{Qualitative results demonstrating multi-attribute transfer using MATTE from a reference image.}
    %\vspace{-10pt}
    \label{fig:main_results_ours}
\end{figure*}

\begin{figure*}[!ht]
    \centering
    \includegraphics[width = 0.9\linewidth]{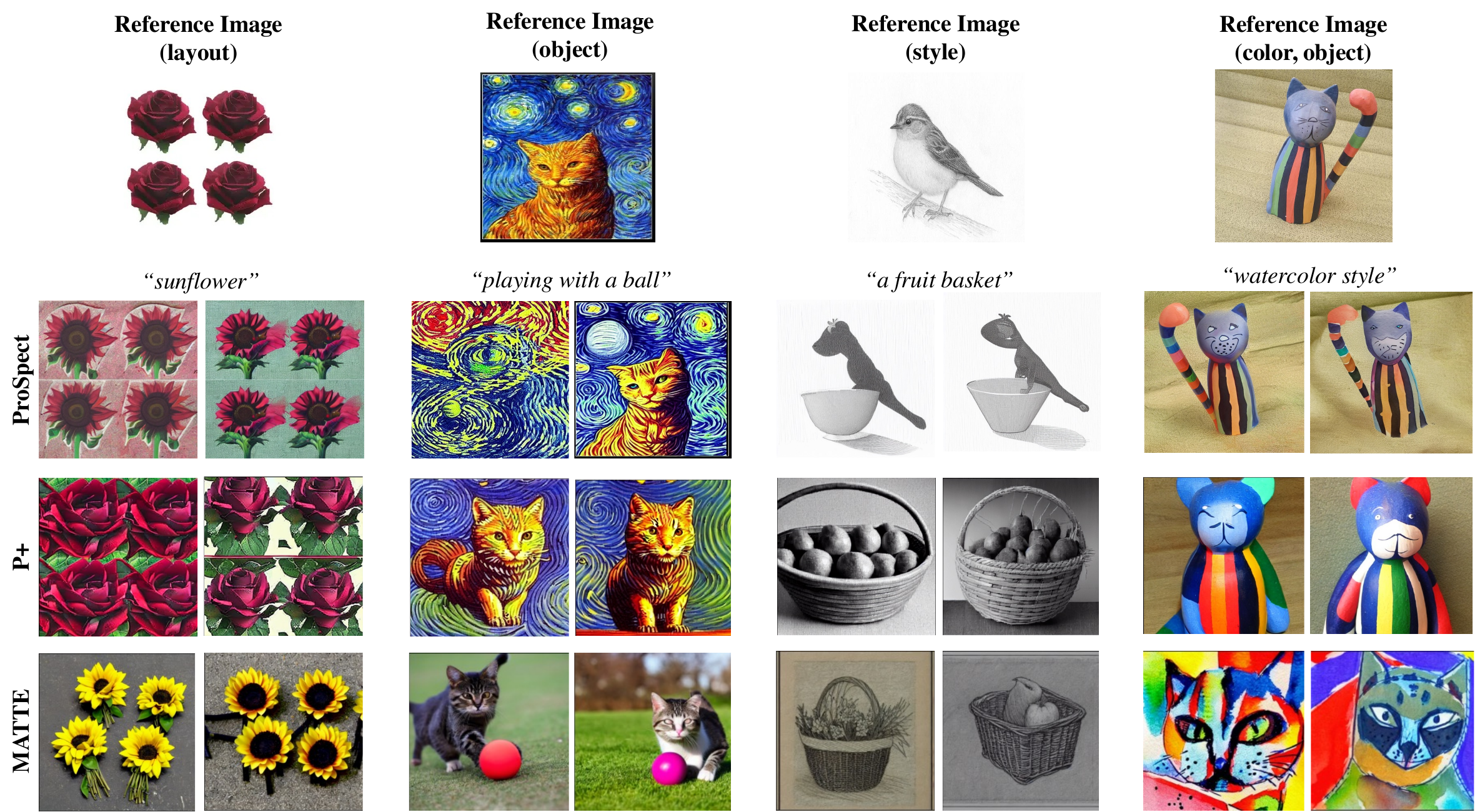}
    %\vspace{-14pt}
    \caption{Comparison of MATTE with recent state-of-the-art methods for reference-constrained text-to-image generation.}
    %\vspace{-10pt}
    \label{fig:qual_comparison}
\end{figure*}

\section{Results}
\label{sec:results}

\textbf{Qualitative Evaluation.} In addition to the results in Figure~\ref{fig:teaser_qual}, we show more results with MATTE in Figure~\ref{fig:main_results_ours} (reference image in first column and our results in columns two-five) to demonstrate multi-attribute transfer from a reference image (all embeddings for $<c>$, $<l>$, $<o>$, $<s>$ are learned with the proposed Equation~\ref{eq:proposed}). In the first row/second column, by specifying $<c>$ and $<o>$ in the input, we are able to generate new cat images in watercolor style following colors of the reference. Similarly, in the last column, we are able to generate new images of a bottle in watercolor style in $<c>$ colors. In the second row/first column, MATTE is able to correctly infer the layout and the object from the reference in synthesizing new images of pebbles stacked on top of each other. Finally, in the third row/third column, MATTE infers the pencil sketch style ($<s>$) of a bird ($<o>$) from the reference to synthesize new images with a prompt comprising both these attributes. %These results show MATTE is learns attributes and conditions the model in synthesizing with individual attributes and novel compositions.

\begin{table}[]
\centering
\scalebox{0.9}{%
\begin{tabular}{@{}c|ccc@{}}
\toprule
Metric                                & $<c>$ & $<o>$ & $<s>$ \\ \midrule 
Cosine Sim. (CLIP Image-Image)         & $0.71$           & $0.72$           & $0.92$          \\ 
Cosine Sim. (CLIP Text-Text)                   & $0.74$ & $0.73$ & $0.87$ \\ \bottomrule
\end{tabular}%
}
    %\vspace{-8pt}
    \caption{CLIP evaluation for tokens learned using MATTE.}
    %\vspace{-4pt}
    \label{tab:token_eval}
\end{table}

\begin{table}[]
\centering
\scalebox{0.6}{%
\begin{tabular}{@{}c|cccccc@{}}
\toprule
Method                             & layout-color & layout-object & layout-style & color-object & color-style & object-style \\ \midrule
P+       &    0.24       &    0.22        &  0.22 &      0.26      &      0.20      &          0.22 \\
ProSpect      &     0.19       &    0.24        & 0.20 &      0.24      &    0.19        &  0.22\\\midrule
\textbf{MATTE }    &    \textbf{0.26 }      &    \textbf{0.27}        &   \textbf{ 0.24 }           &   \textbf{0.28}        &    \textbf{0.26}        &  \textbf{0.26} \\\bottomrule
\end{tabular}%
}
    %\vspace{-8pt}
    \caption{Comparing MATTE with P+ and Prospect.}
    %\vspace{-4pt}
    \label{tab:baseline_comp}
\end{table}

\begin{table}[]
\centering
\scalebox{0.9}{%
\begin{tabular}{@{}c|ccc@{}}
\toprule
Metric                                & $<c>$ & $<o>$ & $<s>$ \\ \midrule
$\mathcal{L_R}$         & $0.62$           & $0.65$           & $0.90$          \\
$\mathcal{L_R} + \mathcal{L_{CS}} + \mathcal{L_O}$                     & \textbf{$0.71$}           & \textbf{$0.72$}           & \textbf{$0.92$} \\ \bottomrule
\end{tabular}%
}
    %\vspace{-8pt}
    \caption{Ablation results for image-image similarities.}
    %\vspace{-4pt}
    \label{tab:token_eval_ablation}
\end{table}

We next compare MATTE with P+ and Prospect in Figure~\ref{fig:qual_comparison}. First row in each example (one example per column) shows the reference and the attribute from it we seek to transfer to synthesized images. The phrase under the reference is part of the prompt used to generate new images (e.g., in first column, the prompt is \texttt{a photo of sunflower following layout $<l>$}, in the fourth column, the prompt is \texttt{a $<c>$ colored photo of $<o>$ in watercolor style}). This is used to generate images with MATTE (shown in the last row in each example). To generate images with P+, we retain the conditioning vectors learned during its inversion process for the layers responsible for the attribute of interest (e.g., in the first column for \texttt{layout}, we retain the set of four \texttt{coarse} prompts \texttt{a photo of sunflower in $<x_i>$}, $i=1,\cdots,4$, where $<x_i>$ is the P+ inverted vector). Similarly, to generate images with Prospect, we retain the vectors learned during its inversion process for timesteps responsible for the corresponding attribute (e.g., for the same \texttt{layout} example, we retain the first three timestep stages prompts). %Please see supplementary details for full implementation details.

In the first column in Figure~\ref{fig:qual_comparison}, one can note that MATTE (last row) is able to respect both the layout from the reference (four items stacked in a two-by-two fashion) as well as the object of interest (\texttt{sunflower}). On the other hand, in P+ (second-last row), since \texttt{object} and \texttt{layout} are captured in the same \texttt{coarse} layers, it is unable to disentangle them and hence we see an unrelated object (rose) even though the layout is respected from the reference image. Similarly, in Prospect, even though the layout from the reference is respected, it produces red-colored sunflowers. This is because the reference is red in color and Prospect is unable to disentangle \texttt{color} from \texttt{layout} since they are both captured in the same timesteps. Similar observations can be made from the other examples. For instance, in the last column, MATTE (last row) is able to successfully transfer both the learned \texttt{color} and the learned \texttt{object} in generating new cat images in watercolor style. In Prospect, \texttt{layout}, \texttt{color}, and \texttt{style} are all captured in the same timesteps. Consequently, as we retain the conditioning vectors learned for \texttt{color} in this example (it is one of the two attributes we wish to transfer), they happen to be entangled with \texttt{layout} and \texttt{style}, resulting in more images that look exactly like the reference image. In P+, we see cat images that follow the reference's \texttt{color} but not in the desired watercolor style because \texttt{color} and \texttt{style} are captured in the same layers, making it impossible to disentangle them. These results provide evidence for our key takeaway message- the four attributes can only be disentangled when optimized \textbf{jointly} across the timestep and layer dimensions with our proposed loss function in Equation~\ref{eq:proposed}. We show additional results, discussion, and limitations in supp.

\textbf{Quantitative Evaluation.} We next quantify the accuracy of the embeddings learned for $<c>$, $<o>$, and $<s>$ from Equation~\ref{eq:proposed} using images from the dataset in \cite{gal2022image}. Note that we are unable to do this for $<l>$ due to the absence of a meaningful ground truth label for layouts. For \texttt{color}, we use Color Thief \cite{dhakar2015color} to extract the ground truth. For \texttt{object} and \texttt{style}, we perform a nearest neighbor lookup using the image's CLIP embedding (see supplementary for more details). For each reference image, we synthesize 64 new images (using \texttt{a $<c>$ colored photo}, \texttt{a $<s>$ style photo}, and \texttt{a photo of $<o>$} with various seeds). We also synthesize a set of corresponding 64 ground-truth images using actual ground-truth labels instead of $<c>$, $<s>$, and $<o>$. Given these two sets, we first compute a CLIP-image-based cosine similarity between the synthesized and ground-truth image. We also compute a CLIP-text-based cosine similarity between the prompts for synthesis (that have $<c>$, $<s>$, and $<o>$ tokens) and the ground-truth prompts (that have ground-truth text labels) (results in Table~\ref{tab:token_eval}). High cosine similarities are indicative of semantic correctness of the learned embeddings for $<c>$, $<s>$, and $<o>$ with our method. We next quantify the improvements with MATTE over P+ and Prospect in Table~\ref{tab:baseline_comp}. For every attribute pair, we evaluate how well methods disentangle them. To do this, in each pair, we keep one of them fixed (e.g., layout) from what is learned during inversion (with all three methods) and vary the other (we have a list of 7 \texttt{style} types, 13 \texttt{object} types, and 11 \texttt{color} types following P+, see supp. for details). In each case, we generate the image and compute the CLIP image-text similarity. A higher score indicates better disentanglement since both attributes would then be separately captured well in the output. As can be seen from Table~\ref{tab:baseline_comp}, this is indeed the case with MATTE outperforming both P+ and Prospect. Finally, we also conduct an ablation study in Table~\ref{tab:token_eval_ablation} where the additional losses $\mathcal{L_{CS}}$ and $\mathcal{L_{O}}$ improve disentanglement by giving higher cosine similarities when compared to $\mathcal{L_{R}}$. 
%generate a similar set of images as described above using tokens $<c>$, $<s>$, and $<o>$ learnt by using just the reconstruction loss. This helps in quantifying the role of the proposed object and content-style disentanglement loss as shown in Table~\ref{tab:token_eval_ablation} \snote{Finish after we have actual numbers}.

\section{Summary}
We presented MATTE, a new algorithm to learn \texttt{color}, \texttt{object}, \texttt{style}, and \texttt{layout} attributes from a reference image and use them for attribute-guided text-to-image synthesis. We first showed existing methods that invert along either the DDPM layer or denoising timestep dimensions are unable to disentangle all the attributes. We then showed that this can be achieved by conditioning both the layer and the timestep dimension as part of the inversion procees, leading to our new inversion algorithm that also comprises explicit disentanglement enhancing regularizers. Extensive evaluations show our method is able to accurately extract attributes from reference image and transfer both individual attributes as well as compositions to new generations.

\begin{appendices}

\section{}
In Section~\ref{subsec:analysis}, we show additional results for joint layer-timestep analysis to show further evidence for how certain attributes can be disentangled when both layer and timestep dimension are considered jointly, which otherwise can not be disentangled across a single dimension (as in P+ \cite{voynov2023p+} and ProSpect \cite{zhang2023prospect}). In Section~\ref{subsec: baselines_setup}, we show more qualitative results for comparing MATTE with baselines. Here we also explain in detail how the prompt conditionings for the baselines P+ and ProSpect are computed. In Section~\ref{subsec: dataset}, we talk about the implementation details and the images used for evaluation. In Section~\ref{subsec:quant_eval_supp}, we provide more details on the quantitative evaluation setup followed for reporting the results comparing MATTE with baselines in Table 2 in the main paper. In Section~\ref{subsec:user_study}, we report results for a user survey conducted to further compare MATTE with baselines. Finally, we conclude with some discussion on
limitations of our method in Section~\ref{subsec:limitations}.

\subsection{Additional Layer-timestep Analysis}
\label{subsec:analysis}
% As discussed in the main paper, an inversion strategy focused on either the layer dimension or the denoising timestep dimension individually is insufficient to disentangle all our attributes of interest from a reference image, and that we need to consider both dimensions \textit{jointly} when designing an inversion strategy so as to learn meaningful individual attribute tokens for subsequent synthesis.
% We had discussed a few results in the main paper which demonstrated that 

As discussed in the main paper,  attributes like \texttt{color} and \texttt{layout} that are captured along the same timesteps (from Prospect \cite{zhang2023prospect}) can be disentangled along the layer dimension. Similarly, geometric attributes like \texttt{object} and \texttt{layout} that are captured along the same layers (from P+ \cite{voynov2023p+}) can be disentangled along the timestep dimension. We show additional qualitative results to demonstrate the conclusions stated above.

Consider Figure~\ref{fig:analysis_layout_color} for an example on layout-color disentanglement using joint layer-timestep prompt conditionings. In Figure~\ref{fig:analysis_layout_color}(a), (b) and (c), one can note that we get a \texttt{blue} ball despite the color being specified as \texttt{red} in the \texttt{coarse} layers. In Figure~\ref{fig:analysis_layout_color}(a) and (b), we get a ball placed \texttt{on a table} and \texttt{under a table} respectively as expected, because the corresponding layout conditionings were given as input to all U-Net \cite{ronneberger2015u} layers. In Figure~\ref{fig:analysis_layout_color}(c), we notice that we get a ball \texttt{on a table} despite specifying \texttt{under a table} in the \texttt{moderate} layers. This clearly indicates that the \texttt{coarse} layers are dominantly responsible for determining the layout. To summarise, this example shows that while \texttt{color} and \texttt{layout} are captured along the same timesteps, they can be disentangled along the layer dimension ($L_3$ - $L_5$ $\&$ $L_{10}$ - $L_{13}$ for color and $L_6$ - $L_9$ for layout).

Similarly, consider the example in Figure~\ref{fig:analysis_layout_object}, where we show \texttt{object} and \texttt{layout} that are captured along the same layers can be disentangled along the timestep dimension. Here, based on the final generated image (a standing cow), one can note that the text conditions corresponding to the object \texttt{cow} were specified in stage $t_2, t_3$ and had the most impact on the final image. For instance, despite the input \texttt{cat} in $t_1$ stage, the final image has a cow. Similarly, despite specifying the layout \texttt{sitting} in stages $t_2, t_3$, the final generation only respected \texttt{standing} that was provided in stage $t_1$. This shows that while \texttt{object} and \texttt{layout} are captured along the same layers, they can be disentangled along the timestep dimension ($t_2, t_3$ for object and $t_1$ for layout).

Before we move on to the next example, we had summarised from our analysis in the main paper that fine layers ($L_1$ - $L_2$ $\&$ $L_{14}$ - $L_{16}$) and the stage $t_4$ have no impact on any of the four attributes. \texttt{Color} and \texttt{style} are both captured in the initial denoising stages ($t_1, t_2$) and across moderate U-Net layers ($L_3$ - $L_5$ $\&$ $L_{10}$ - $L_{13}$). \texttt{Object} semantics are captured along the middle denoising stages ($t_2, t_3$) and across the coarse U-Net layers ($L_6$ - $L_9$). Finally, \texttt{layout} is captured in the very initial denoising stage ($t_1$) across coarse layers ($L_6$ - $L_9$).

On that note, we show another example in Figure ~\ref{fig:analysis_all_mega}, similar to the joint multi-prompt conditioning example presented in Figure 5 in the main paper, which demonstrates the properties summarised above. For the final generated image (\textit{a blue ball placed on a table}), one can note that the textual conditionings corresponding to each of key attributes in the prompt (\texttt{blue}, \texttt{ball}, and \texttt{on the table}) were specified only across a subset of layers and only along specific timesteps. These observations are consistent with the analysis we summarised for each of the attributes. For instance, , despite specifying \texttt{white} in $L_1$ - $L_2$ $\&$ $L_{14}$ - $L_{16}$ layers, and color \texttt{red} in ($L_6$ - $L_9$) layers,  we still see a \texttt{blue} colored ball based on the color specified in moderate ($L_3$ - $L_5$ $\&$ $L_{10}$ - $L_{13}$) layers. Similarly, the layout is captured in the initial stages and across the coarse set of layers. We also show per-layer attention maps across denoising timesteps in Figure~\ref{fig:analysis_layout_ca_maps} which confirms that layout is predominantly captured in layers $L_6$, $L_8$, and $L_9$.

\begin{figure*}[h]
    \centering
    \includegraphics[width = \linewidth]{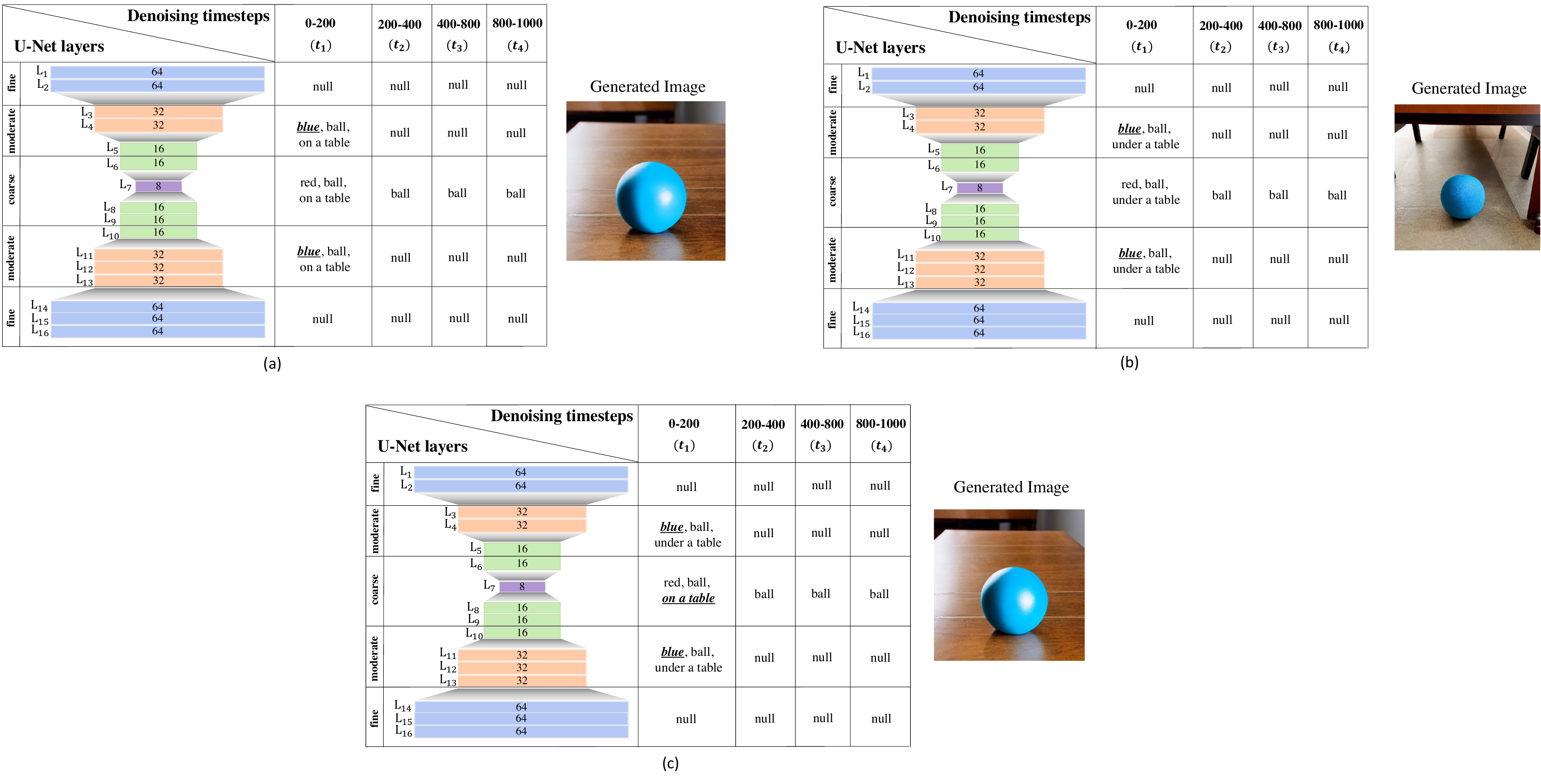}
    %\vspace{-18pt}
    \caption{Layout-Color disentanglement.}
    %\vspace{-5pt}
    \label{fig:analysis_layout_color}
\end{figure*}

\begin{figure*}[h]
    \centering
    \includegraphics[width = 0.85\linewidth]{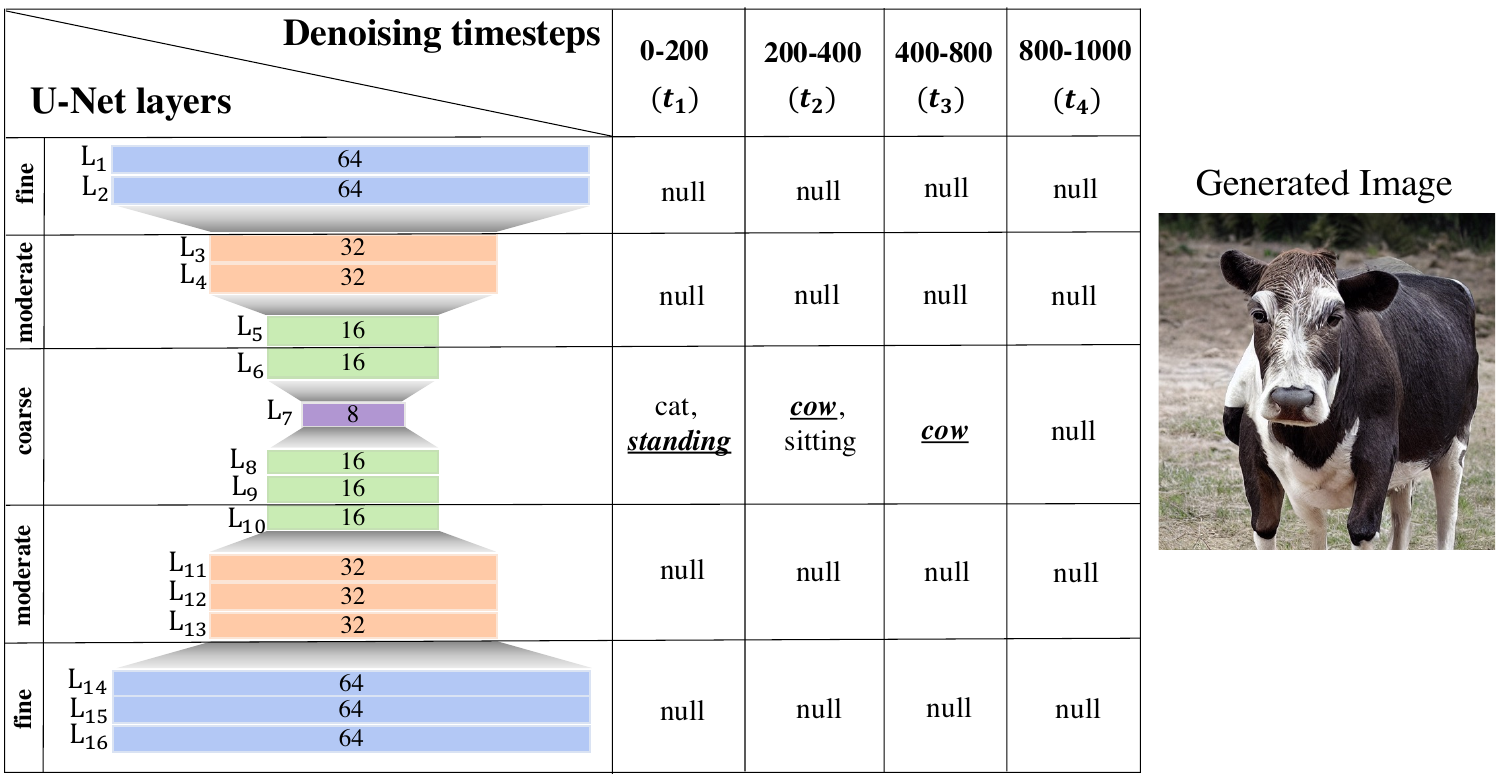}
    %\vspace{-18pt}
    \caption{Layout-Object disentanglement.}
    %\vspace{-5pt}
    \label{fig:analysis_layout_object}
\end{figure*}

% \begin{figure}[h]
%     \centering
%     \includegraphics[width = \linewidth]{figs/cl_ball_1.pdf}
%     %\vspace{-18pt}
%     \caption{Layout-Color disentanglement.}
%     %\vspace{-5pt}
%     \label{fig:analysis_layout_color_1}
% \end{figure}

% \begin{figure}[h]
%     \centering
%     \includegraphics[width = \linewidth]{figs/cl_2.pdf}
%     %\vspace{-18pt}
%     \caption{Layout-Color disentanglement.}
%     %\vspace{-5pt}
%     \label{fig:analysis_layout_color_2}
% \end{figure}

% \begin{figure}[h]
%     \centering
%     \includegraphics[width = \linewidth]{figs/cl_ball_3.pdf}
%     %\vspace{-18pt}
%     \caption{Layout-Color disentanglement.}
%     %\vspace{-5pt}
%     \label{fig:analysis_layout_color_3}
% \end{figure}

\begin{figure*}[h]
    \centering
    \includegraphics[width = \linewidth]{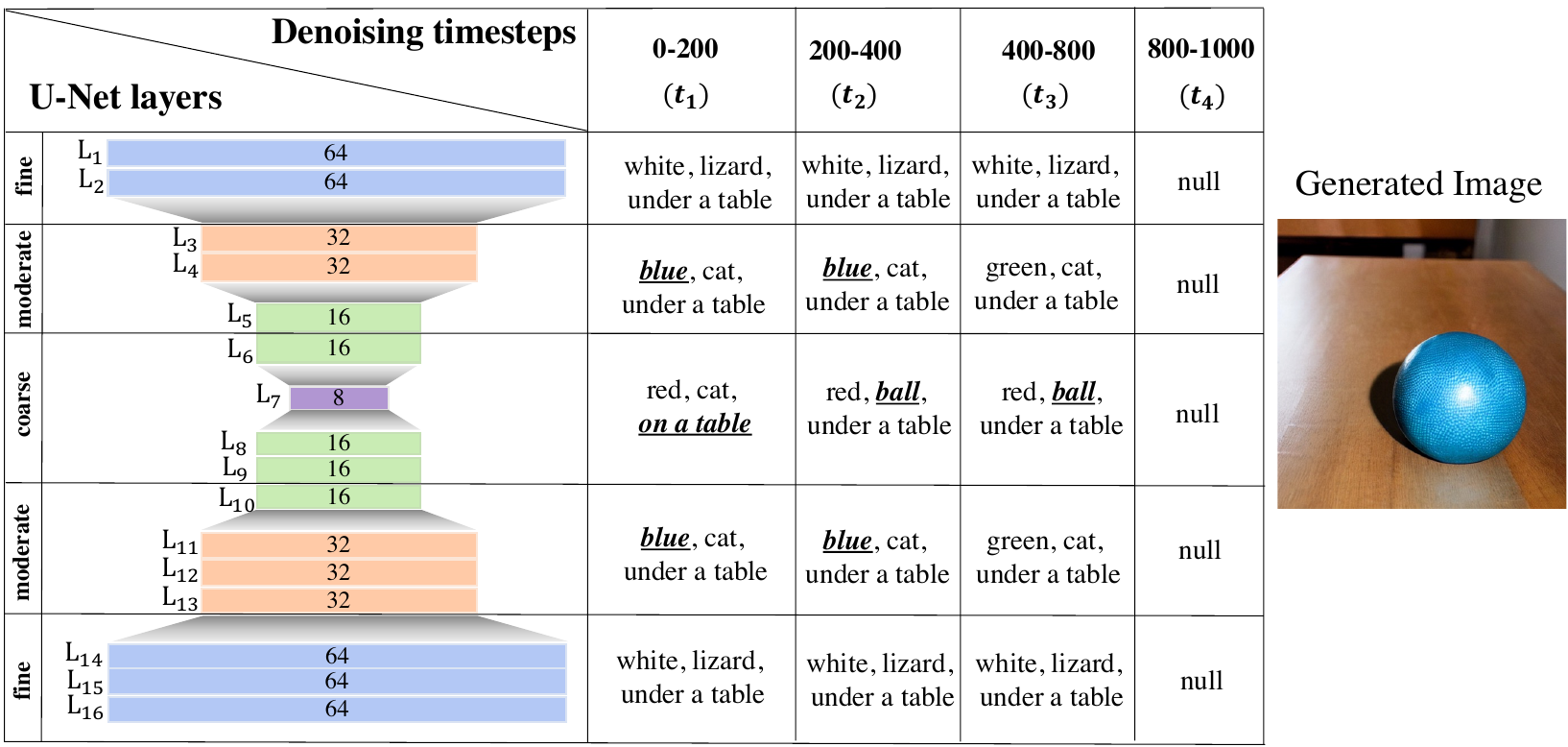}
    %\vspace{-18pt}
    \caption{Multi-prompt conditioning across U-Net layers and denoising timesteps jointly.}
    %\vspace{-5pt}
    \label{fig:analysis_all_mega}
\end{figure*}

\begin{figure}[h]
    \centering
    \includegraphics[width = \linewidth]{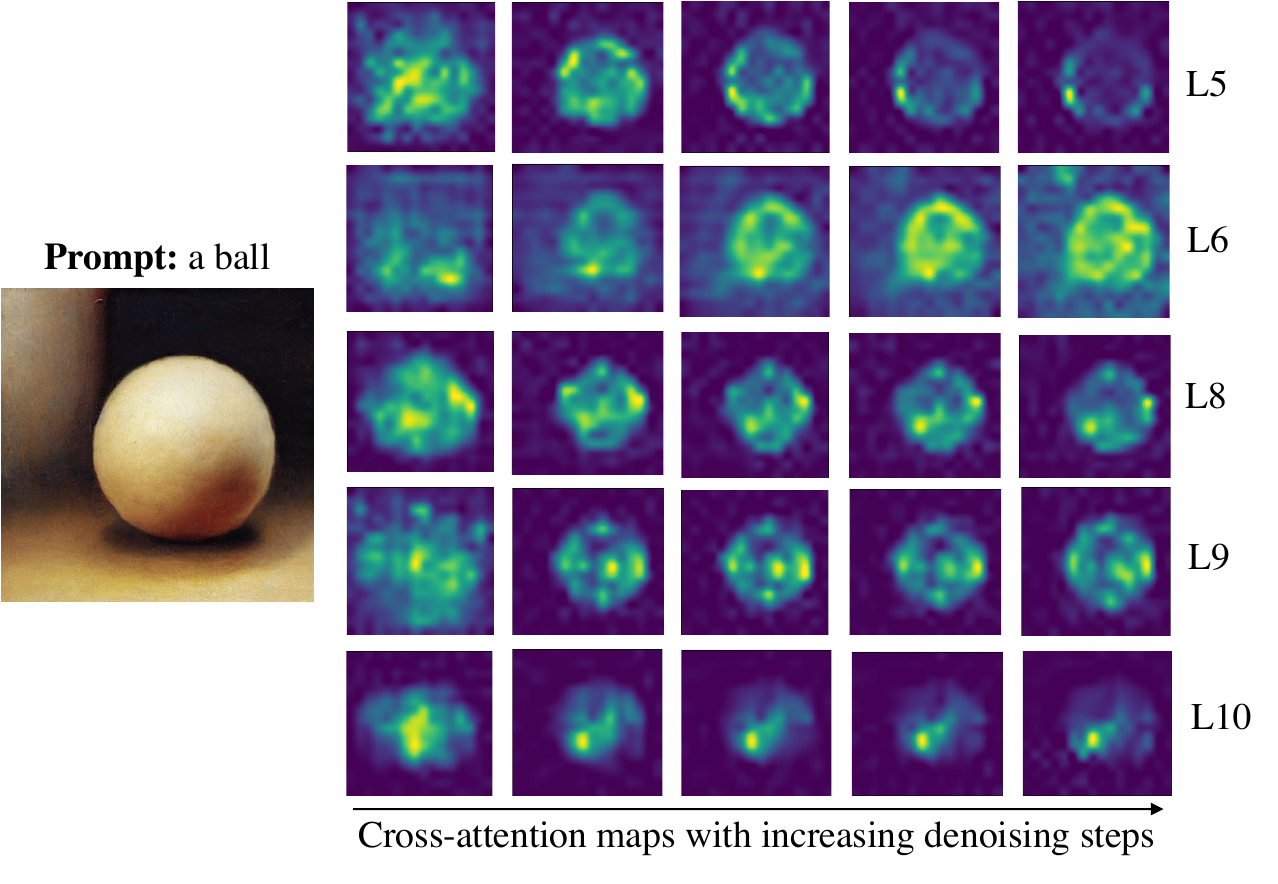}
    %\vspace{-18pt}
    \caption{Cross-attention maps for analysing layout.}
    %\vspace{-5pt}
    \label{fig:analysis_layout_ca_maps}
\end{figure}

\subsection{Additional Qualitative Results and Setup Details}
\label{subsec: baselines_setup}

\begin{figure*}[h]
    \centering
    \includegraphics[width = \linewidth]{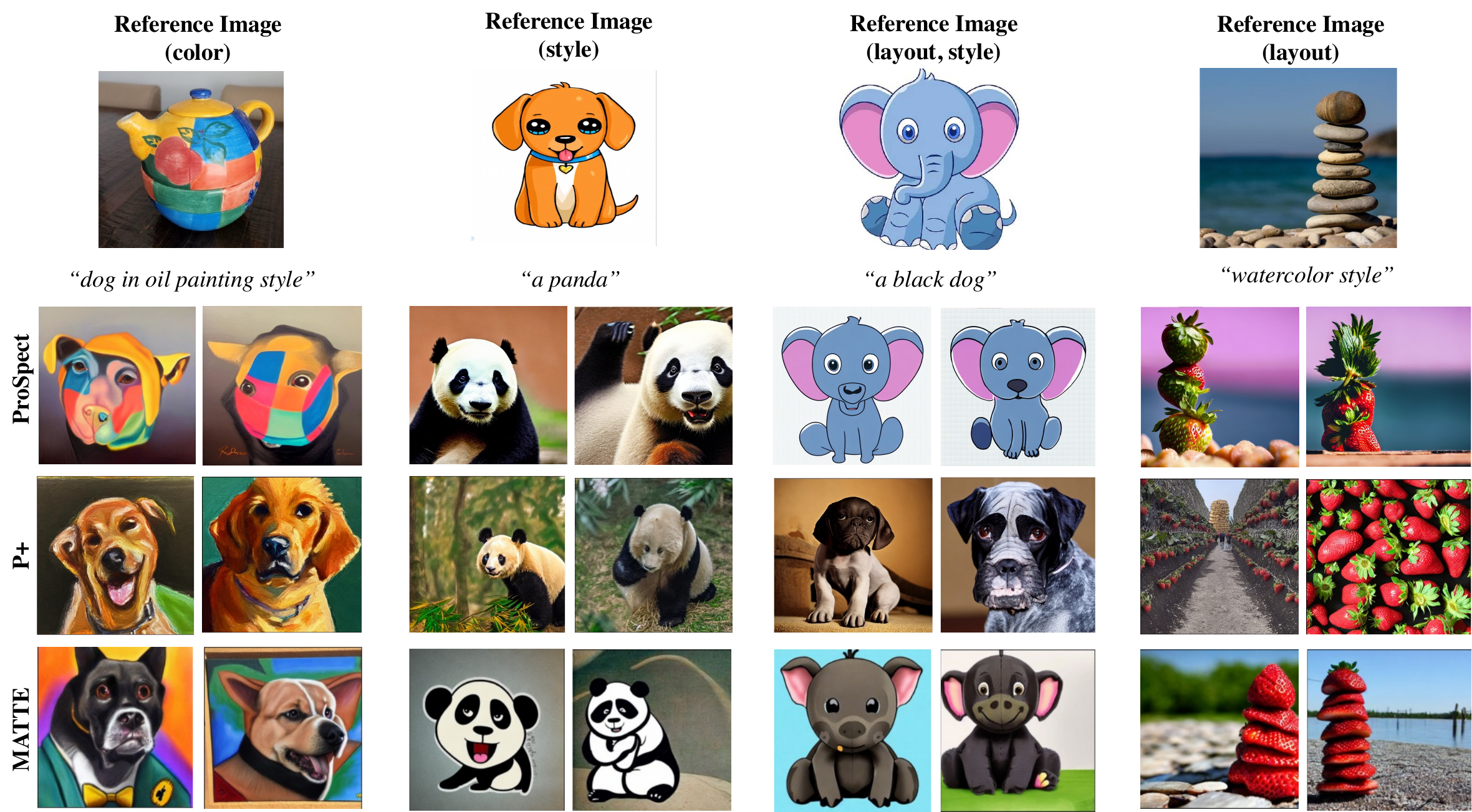}
    %\vspace{-18pt}
    \caption{Comparison of MATTE with recent state-of-the-art methods for reference-constrained text-to-image generation.}
    %\vspace{-5pt}
    \label{fig:baseline_comp_supp}
\end{figure*}

We show additional results comparing MATTE with the closest baselines ProSpect \cite{zhang2023prospect} and P+ \cite{voynov2023p+}.
We first explain how we generate images using P+ and ProSpect given a prompt with an example. Consider the example shown in column 1 in Figure~\ref{fig:baseline_comp_supp}. The goal here is to generate images of a \texttt{dog} in \texttt{oil painting} style following the \texttt{color} properties of the reference image. The first step here is to run the inversion algorithms of P+ and ProSpect for the reference image, and get a set of textual conditionings $<x_i>$ where $i=1,\cdots,16$ for P+, and $<y_j>$ where $j=1,\cdots,10$ for ProSpect.
Next, depending upon the attributes we want to transfer from the reference image (\texttt{color} here), we retain the textual conditionings learned during inversion in P+ and Prospect as part of the final conditionings used as input along the $16$ layers and $10$ timestep stages respectively. The decision of retaining conditionings is made on the basis of which set of timesteps/layers are important for capturing the attribute of interest (color here). Since we know color is captured across the shallow U-Net layers in P+, and across the initial denoising stages in ProSpect, the prompt that goes as input to P+ across the $16$ U-Net layers is:\\
\texttt{[$<x_1>$ dog in oil painting style,\\ $<x_2>$ dog in oil painting style,\\ $<x_3>$ dog in oil painting style,\\ $<x_4>$ dog in oil painting style,\\ $<x_5>$ dog in oil painting style,\\ dog,\\ dog,\\ dog,\\ dog,\\ $<x_{10}>$ dog in oil painting style,\\ $<x_{11}>$ dog in oil painting style,\\ $<x_{12}>$ dog in oil painting style,\\ $<x_{13}>$ dog in oil painting style,\\ $<x_{14}>$ dog in oil painting style,\\ $<x_{15}>$ dog in oil painting style,\\ $<x_{16}>$ dog in oil painting style]}.\\

Similarly, the prompt for ProSpect across the $10$ denoinsing timestep stages is:\\
\texttt{[$<y_1>$ dog in oil painting style,\\ $<y_2>$ dog in oil painting style,\\ $<y_3>$ dog in oil painting style,\\ $<y_4>$ dog in oil painting style,\\ dog in oil painting style,\\ dog in oil painting style,\\ dog in oil painting style,\\ dog in oil painting style,\\ dog in oil painting style,\\ dog in oil painting style]}.\\

We next discuss the results comparing MATTE with P+ and ProSpect in Figure~\ref{fig:baseline_comp_supp}.
Consider the example in column 1. Here the goal is to generate a \texttt{dog} in \texttt{oil painting style} while retaining only the \texttt{color} properties from the reference image. We see that MATTE captures everthing (dog, oil painting style and color attribute from reference image) accurately. While in ProSpect, even though the colors got transferred from the reference image, but it has generated dogs following the layout of the inkpot shown in the reference image. This is because, as seen previously, layout and color are captured across similar denoising timesteps, hence disentangling the two is not possible in ProSpect (as inversion in ProSpect is across timestep dimension only). Similarly, for P+, we see that the generated dogs follow the oil painting style but are unable to capture the color of the inkpot. This again is because color and style are captured in same layers in P+, so either color and style both get transferred together or none gets transferred. One can make similar observations across the examples shown in other columns too which clearly indicated that MATTE is able to constrain the generation of images on attributes from reference image in a disentangled fashion much better than the closest baselines.

\subsection{Implementation Details and Dataset}
\label{subsec: dataset}

We follow the same set of styles, objects and colors as described in P+ \cite{voynov2023p+} for all our evaluations and trainings. 

Specifically, during MATTE inversion technique (Section 3.2 in the main paper), the set of styles used to randomly choose styles from was:\\
\texttt{["oil painting", "vector art", "pop art style", "3D rendering", "impressionism picture", "graffiti", "fuzzy", "shiny", "bright", "fluffy", "sparkly", "dull", "smooth", "rough", "jagged", "striped", "painting", "retro", "vintage", "modern", "bohemian", "industrial", "rustic", "classic", "contemporary", "futuristic"]}

For the quantitative evaluations in Section 4 in the main paper, we use the following sets of objects, style and colors (again from P+ \cite{voynov2023p+}):\\

Objects = \texttt{["chair", "dog", "book", "elephant", "guitar", "pillow", "rabbit", "umbrella", "yacht", "house", "cube", "sphere", "car’"]}\\

Colors = \texttt{["black", "blue", "brown", "gray", "green", "orange", "pink", "purple", "red", "white", "yellow"]}\\

Styles = \texttt{["watercolor", "oil painting", "vector art", "pop art style", "3D rendering", "impressionism picture", "graffiti"]}\\

Finally, we show the images used for different evaluation setups in Figure~\ref{fig:data_images_eval}.

\subsection{Quantitative Evaluation Setup Details}
\label{subsec:quant_eval_supp}
\begin{figure*}[h]
    \centering
    \includegraphics[width = \linewidth]{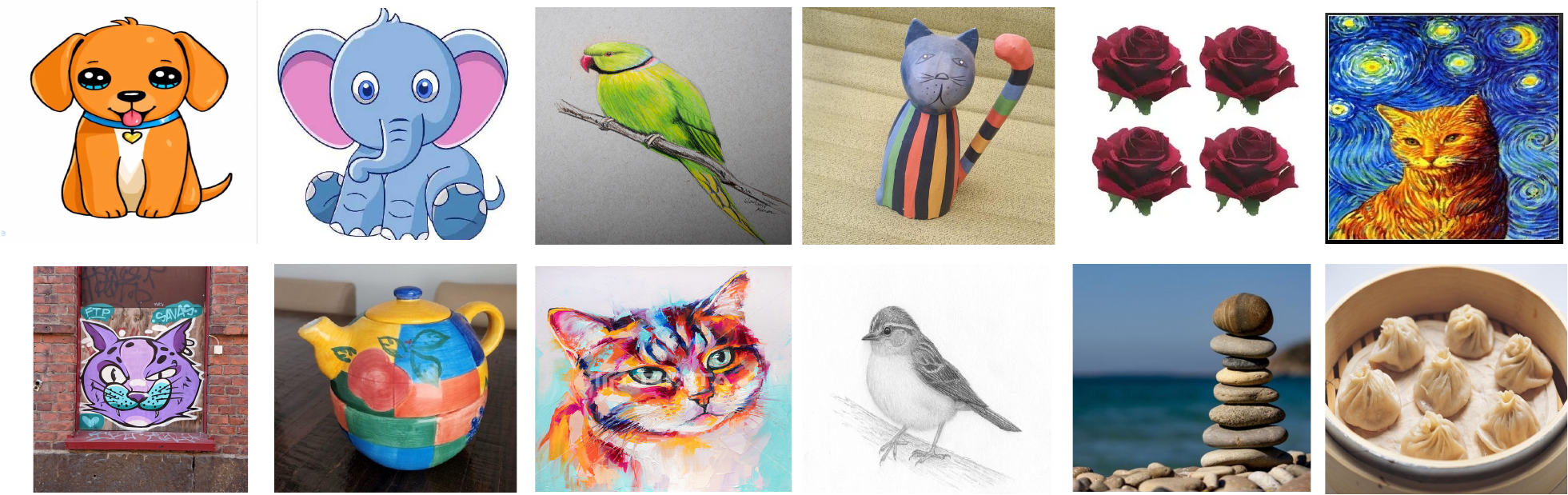}
    %\vspace{-18pt}
    \caption{Images used for evaluation.}
    %\vspace{-5pt}
    \label{fig:data_images_eval}
\end{figure*}

We presented an evaluation to quantify the disentanglement of different pairs of attributes in the main paper in Table 2, Section 4. Here, we explain the details of how we compute the CLIP Image-text similarities reported in the paper. We use the set of images shown in Figure~\ref{fig:data_images_eval} and the set of attributes discussed in Section~\ref{subsec: dataset}. 
For each reference image, our goal was to evaluate the inversion techniques in aspects of (i) preserving/transferring an attribute from the reference image and (ii) generating images following attributes mentioned in the text prompt. 
We considered $6$ unique pairs of attributes for this comparison namely \texttt{layout-color, layout-object, layout-style, color-object, color-style} and \texttt{object-style}. 
Consider the case of \texttt{color-object} disentanglement evaluation in the context of reference image based attribute-aware text-to-image generation. The attribute mentioned first (\texttt{color} here) is the one to be transferred from the reference image, whereas the latter (\texttt{object} here) comes from the text prompt. For each of the baselines P+ and ProSpect, we generate final prompt conditionings in the same fashion as explained in Section~\ref{subsec: baselines_setup} by retaining the textual conditionings responsible for capturing the attribute to be transferred from reference image (color here). For the attribute that comes from the text prompt (objects here), we iterate over a set of different objects following the list of objects mentioned in Section~\ref{subsec: dataset} and generate a set of $64$ images for each color-object pair. We then compute CLIP Image-text similarities between the generated images and the ground truth object used to generate those images. Similarly, we also compute CLIP Image-text similarities between the generated images and the corresponding ground truth for the attribute to be transferred from reference image wherever possible, followed by an averaging of the two similarities (for color in color-object case, ground truth colors are extracted from the reference image using Color Thief \cite{dhakar2015color}). Similarly, these CLIP based Image-text similarities are computed for other attribute pairs for MATTE and the closest baselines P+ and ProSpect, results of which are reported in Table 2 in the main paper.

\subsection{User Study}
\label{subsec:user_study}
\begin{figure}[h]
    \centering
    \includegraphics[width = \linewidth]{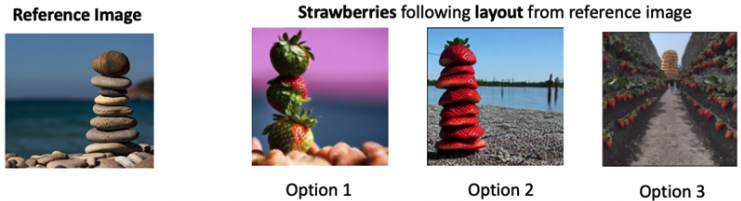}
    %\vspace{-18pt}
    \caption{A sample question from the conducted user study.}
    %\vspace{-5pt}
    \label{fig:user_study_example}
\end{figure}

We conduct a user study with the generated images where we ask survey respondents to select which set of images (among sets from three different methods, see Table~\ref{tab:userStudy}) best represents the input constraints. The user is presented with a reference image, a text prompt, and a set of attributes from the reference image that should ideally get transferred to the final generated image (see Figure~\ref{fig:user_study_example} for an example). From Table~\ref{tab:userStudy}, our method's results are preferred by a majority of the survey respondents, thus providing additional evidence for the impact of our proposed inversion technique in constraining text-to-image generation on different attributes of reference images in a disentangled fashion.

\begin{table}[]
\centering
\resizebox{0.35\textwidth}{!}{%
\begin{tabular}{@{}c|c@{}}
\toprule
Method                                & Percentage \\ \midrule
P+ \cite{voynov2023p+}     & $12.2\%$              \\
ProSpect \cite{zhang2023prospect} & $13.2\%$               \\ \midrule
\textbf{MATTE}                     & $\textbf{74.6\%}$  \\ \bottomrule
\end{tabular}%
}
    % \vspace{-8pt}
    \caption{Results from a user survey with $24$ respondents.}
    % \vspace{-4pt}
    \label{tab:userStudy}
\end{table}

\subsection{Limitations}
\label{subsec:limitations}

In this Section, we briefly discuss a few limitations of MATTE when seen in a constrained text-to-image generation setup. Firstly, the optimization of the embeddings learned for the four tokens namely $<c>$, $<l>$, $<o>$ and $<s>$ during inversion is a slow process (MATTE converges faster than TI \cite{gal2022image}, but is still slow), thereby posing a limitation to its' practical applicability. Secondly, since MATTE doesn't involve fine-tuning model weights, the final constrained text-to-image generation pipeline after MATTE inverts the reference image into disentangled tokens is limited by the generation abilities of the base diffusion model. For instance, omission of objects mentioned in the text prompt is a known limitation of diffusion models \cite{agarwal2023star, chefer2023attend, feng2022training, liu2022compositional}. So, given a prompt \texttt{"a <c> colored cat playing with a dog"} (where $<c>$ is extracted from a reference image using MATTE) to the base diffusion model, MATTE will ensure that the cat generated is $<c>$ colored but MATTE can not enforce the presence of a cat in the final generated image.

\end{appendices}

\bibliography{aaai24}

\end{document}